\documentclass[10pt,twocolumn,letterpaper]{article}

\usepackage{cvpr}              

\usepackage{lipsum}
\usepackage{graphicx}
\usepackage{placeins}
\usepackage{multirow}
\definecolor{cvprblue}{rgb}{0.21,0.49,0.74}
\usepackage[pagebackref,breaklinks,colorlinks,allcolors=cvprblue]{hyperref}

\newcommand{\paragrapht}[1]{\noindent\textbf{#1}}

\newcommand{\equalcontributionnotice}{%
    \begingroup
    \renewcommand{\thefootnote}{\fnsymbol{footnote}}%
    \footnotetext[1]{Equal contribution.}%
    \endgroup
}

\def\paperID{} 
\def\confName{CVPR}
\def\confYear{2026}

\title{Attribute-Preserving Pseudo-Labeling for Diffusion-Based Face Swapping}

\author{
    Jiwon Kang\textsuperscript{*1} \quad
    Yeji Choi\textsuperscript{*1} \quad
    JoungBin Lee\textsuperscript{1} \quad
    Wooseok Jang\textsuperscript{1} \\
    Jinhyeok Choi\textsuperscript{1} \quad
    Taekeun Kang\textsuperscript{2} \quad
    Yongjae Park\textsuperscript{2} \quad
    Myungin Kim\textsuperscript{2} \quad
    Seungryong Kim\textsuperscript{1}\\[3pt] 
    \textsuperscript{1}KAIST AI \quad \textsuperscript{2}SAMSUNG\\[3pt]
{\tt \href{https://cvlab-kaist.github.io/APPLE}{https://cvlab-kaist.github.io/APPLE}}
}

\usepackage[toc,page,header]{appendix}
\usepackage{minitoc}

\usepackage{xspace}
\newcommand{\ourframework}{APPLE\xspace}

\begin{document}

\twocolumn[{
\maketitle

\begin{center}
\vspace{-15pt}

\includegraphics[width=\linewidth]{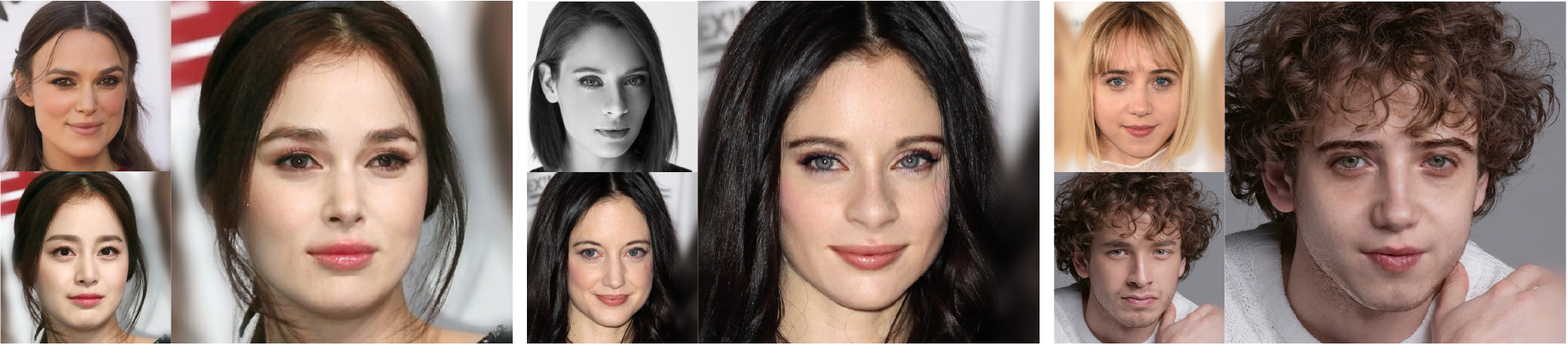} 
    \captionof{figure}{ \textbf{\ourframework} (\textbf{A}ttribute-\textbf{P}reserving \textbf{P}seudo-\textbf{L}ab\textbf{e}ling) successfully transfers the identity of a source (\textbf{top left}) onto a target (\textbf{bottom left}) while accurately preserving target attributes (e.g. pose, expression, skin tone, lighting) across ethnicity, input variations, and gender.
    }
\label{fig:teaser}
\end{center}
}] 
\equalcontributionnotice

\begin{abstract}

\vspace{-11pt}


Face swapping aims to transfer the identity of a source face onto a target face while preserving target-specific attributes such as pose, expression, lighting, skin tone, and makeup. However, since real ground truth for face swapping is unavailable, achieving both accurate identity transfer and high-quality attribute preservation remains challenging. Recent diffusion-based approaches attempt to improve visual fidelity through conditional inpainting on masked target images, but the masked condition removes crucial appearance cues, resulting in plausible yet misaligned attributes. To address this limitation, we propose \textbf{\ourframework} (\textbf{A}ttribute-\textbf{P}reserving \textbf{P}seudo-\textbf{L}ab\textbf{e}ling), a fully diffusion-based teacher–student framework for attribute-preserving face swapping. Our approach introduces a teacher design to produce pseudo-labels aligned with the target attributes through (1) a conditional deblurring formulation that improves the preservation of global attributes such as skin tone and illumination, and (2) an attribute-aware inversion scheme that further enhances fine-grained attribute preservation such as makeup. \ourframework conditions the student on clean pseudo-labels rather than degraded masked inputs, enabling more faithful attribute preservation. As a result, \ourframework achieves state-of-the-art performance in attribute preservation while maintaining competitive identity transferability.


\end{abstract}    
\section{Introduction}
\label{sec:intro}

Face swapping aims to replace the identity of a person in a target image with that of a source image while faithfully preserving target-specific attributes such as pose, expression, skin tone, lighting, gaze, makeup, and accessories. It is widely applied in digital content creation, privacy protection, and film production, emphasizing the importance of generating high-quality outputs. However, since real face swapping data do not exist, supervised training of face swapping models is fundamentally infeasible, making it challenging to ensure accurate identity transfer and consistent attribute preservation at high fidelity.

Early face swapping approaches~\citep{chen2020simswap, zhu2021one_megafs, wang2021hififace, gao2021information_infoswap, yuan2023reliableswap} primarily adopt GAN-based methods~\cite{goodfellow2014generative}, leveraging the generative capability of GANs to perform identity replacement through facial editing. However, these models typically rely on two conflicting objectives: an identity loss that enforces source identity transfer, and reconstruction or attribute-related losses that encourage target attribute preservation. Such indirect and competing supervision makes GAN training unstable and requires extensive hyperparameter tuning, often resulting in copy-and-paste-like artifacts and visually unnatural outputs~\cite{ye2025dreamid,shao2025vividface}.

Recently, the strong generative priors of diffusion models \citep{ho2020ddpm, rombach2022high, esser2024scaling, flux2024} have driven significant progress in various image synthesis tasks, providing high-quality synthesis, precise conditional control, and reliable training stability. Motivated by these advantages, diffusion-based face-swapping methods \citep{zhao2023diffswap, han2024faceadapter, yu2025reface} formulate the task as conditional inpainting. Under this proxy training objective, the model is trained to reconstruct the original clean target image from its masked version, given a source image of the same identity but with different attributes. However, this masked conditioning removes crucial cues from the target, such as lighting, skin tone, makeup, and subtle expression dynamics. Consequently, even with auxiliary attribute information like 3DMM \citep{blanz2023morphable_3dmm} landmarks or CLIP \citep{radford2021learning} features, existing models often fail to preserve target-specific attributes during face swapping.


To address the issue of misaligned attributes, we propose \textbf{\ourframework} (\textbf{A}ttribute-\textbf{P}reserving \textbf{P}seudo-\textbf{L}ab\textbf{e}ling), a fully diffusion-based teacher–student framework for attribute-preserving face swapping. The core idea is to train a teacher model to generate high-quality pseudo-labels aligned with the target’s visual attributes and use them as conditioning inputs for the student. Unlike conventional inpainting-based methods that rely on masked conditions, the student is conditioned on clean, unmasked pseudo-labels that provide richer attribute cues and is supervised under a direct image-editing objective to reconstruct the original target image, thereby achieving high-fidelity attribute preservation.


\begin{figure}[t]
    \centering
    \includegraphics[width=\linewidth]{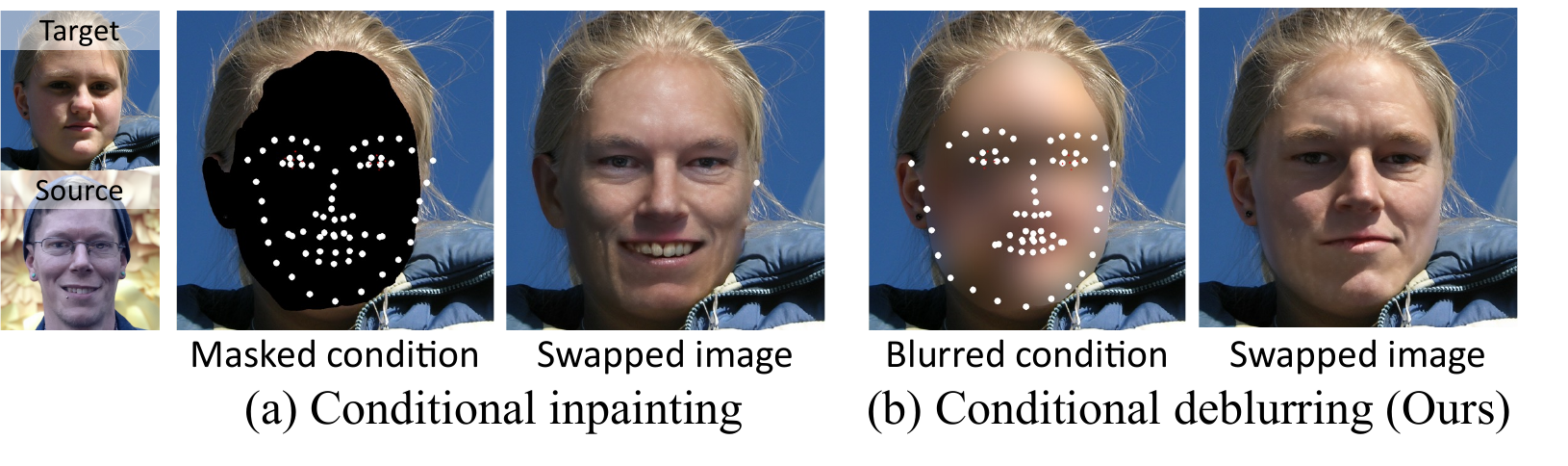}
    \vspace{-15pt}
    \captionof{figure}{
        \textbf{Comparison of conditioning methods.} Compared to conditional inpainting widely used in existing works~\citep{zhao2023diffswap,yu2025reface,han2024faceadapter}, the proposed conditional deblurring strategy achieves largely improved attribute (e.g., lighting) preservation of targets.  }
    \label{fig:condition_comparison}
    \vspace{-15pt}
\end{figure}

Importantly, the effectiveness of this training scheme depends on the teacher’s ability to generate pseudo-labels that maintain attribute consistency with the target image. Without the consistency, the student receives conflicting signals (e.g., in pose or lighting), which degrades attribute preservation.
To this end, APPLE first replaces the conventional conditional inpainting objective of the teacher diffusion model with a conditional deblurring formulation, leading to improved preservation of attributes such as skin tone and illumination. Second, to further enhance fine-grained attribute preservation during pseudo-label generation, we introduce an attribute-aware inversion scheme that intentionally exploits the reduced editability of inverted noise to anchor fine-grained details. Together, these designs enable the teacher to produce attribute-aligned pseudo-labels, allowing the student to learn under a direct editing objective and achieve reliable identity transfer with improved preservation of fine-grained target attributes.

In summary, APPLE achieves state-of-the-art face-swapping performance in terms of attribute preservation while maintaining competitive identity transferability, yielding more coherent and photorealistic images than previous methods. Our main contributions are as follows:

\begin{itemize}

\item We propose \textbf{\ourframework}, a diffusion-based teacher–student framework that leverages improved pseudo-label quality as the key to achieving superior attribute preservation.

\item We enhance the attribute preservation of the teacher's output by replacing the conventional conditional inpainting formulation with a \textit{conditional deblurring} objective during training and introducing an \textit{attribute-aware inversion} scheme during inference.

\item We demonstrate that by training on these high-fidelity pseudo-labels, the student model achieves state-of-the-art attribute-preservation while maintaining high identity similarity.

\item Our framework offers high practical value for real-world deployment, as the student model requires no complex preprocessing for attribute conditioning at inference.

\end{itemize}

\begin{figure*}[t!] 
    \centering
    \includegraphics[width=\textwidth]{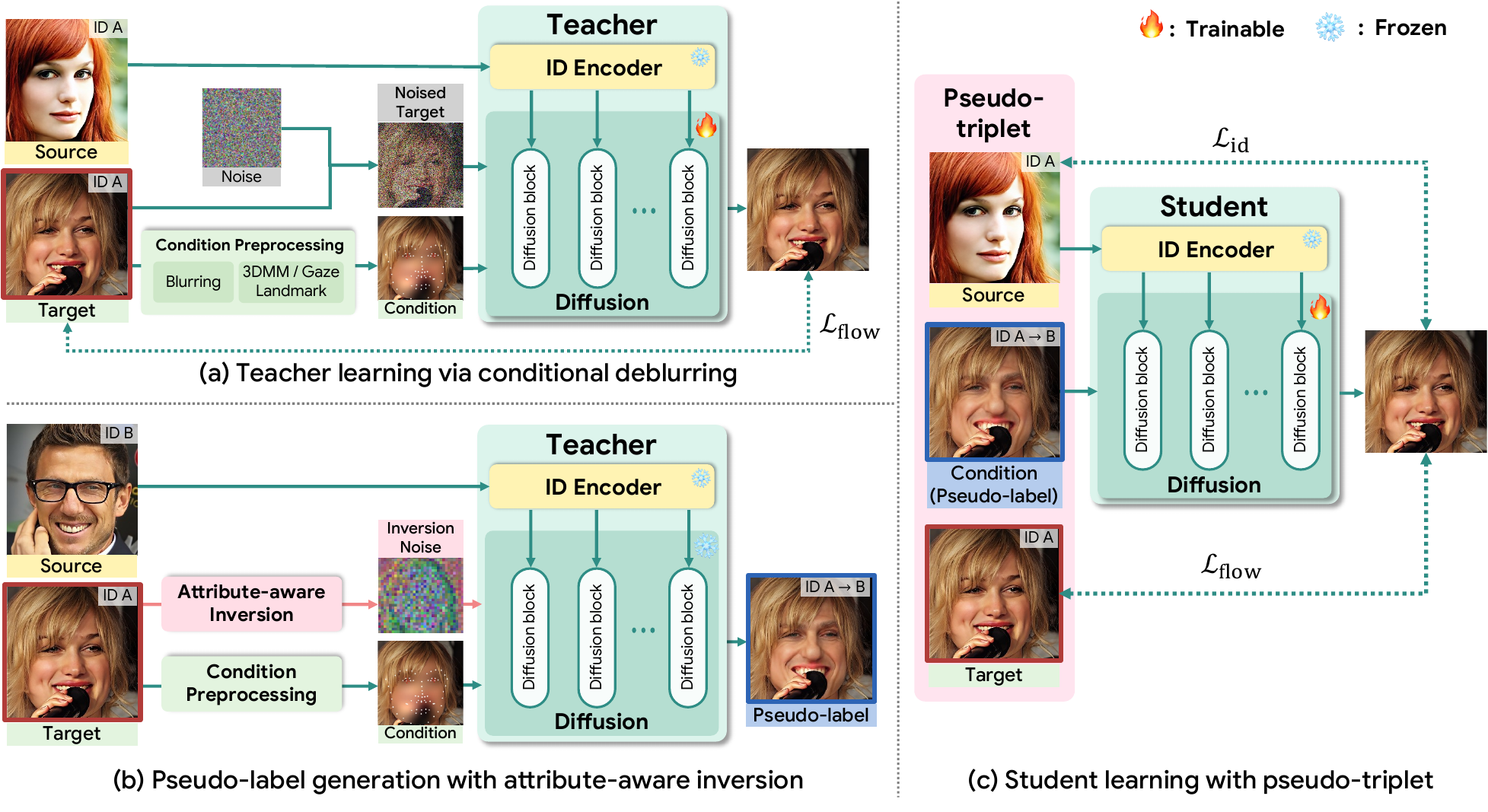}
    \vspace{-20pt}
    \caption{\textbf{Overall architecture of the proposed method.} We propose \textbf{\ourframework}, a diffusion-based teacher-student framework that focuses on improving attribute preservation.   (a) To improve target-attribute preservation, we propose training a teacher with conditional deblurring rather than conditional inpainting widely used in existing works~\citep{zhao2023diffswap,yu2025reface,han2024faceadapter}. (b) When constructing a pseudo-label with a teacher, we propose \textit{attribute-aware inversion} which further improves fine-grained attribute preservation in inference time. Note that inversion noise cannot be used during training due to its non-Gaussian property. (c) The student model is trained using attribute-aligned pseudo-labels generated by the teacher. By leveraging high-fidelity attribute conditioning—clean images rather than degraded inputs—the student model eventually outperforms the teacher, achieving state-of-the-art attribute-preservation while maintaining high identity similarity. Note that the noised target input to the diffusion model is omitted in (c).
}
    \label{fig:overview}
\end{figure*}

\section{Related Work}
\label{sec:relwork}

\paragraph{GAN-based models.}
GANs~\cite{goodfellow2014generative} play a dominant role in early face-swapping research~\cite{nirkin2019fsgan,li2019faceshifter, gao2021information_infoswap, chen2020simswap}. 
FSGAN~\cite{nirkin2019fsgan} introduces an identity-agnostic framework that fuses source identity with target attributes through adaptive blending. SimSwap~\cite{chen2020simswap} proposes an ID-injection module that embeds identity information and applies a weak feature-matching loss to preserve target attributes, while SimSwap++~\cite{chen2023simswap++} further enhances the model’s efficiency. 
To improve robustness under pose and expression variations, HiFiFace~\cite{wang2021hififace} incorporates 3D priors (e.g., 3DMM~\cite{blanz2023morphable_3dmm}) and face-recognition constraints to achieve geometrically consistent face generation.
In parallel, StyleGAN-based models~\cite{karras2019style} are widely adopted for their high-fidelity generative capability.
Representative works, such as FaceDancer~\cite{rosberg2023facedancer} and E4S~\cite{liu2023fine_e4s}, further enhance visual realism by employing adaptive feature fusion attention for hierarchical feature integration and region-based inversion within the StyleGAN latent space.
On another branch, to compensate for the absence of ground-truth supervision, recent methods such as CSCS~\cite{huang2024identity_cscs} and ReliableSwap~\cite{yuan2023reliableswap} generate pseudo pairs using pre-trained GAN models to augment training data and improve identity consistency.
However, such pseudo samples often suffer from attribute misalignment and visual artifacts inherited from the GAN generator, limiting their effectiveness as reliable supervision.

Overall, GAN-based approaches still rely on complex loss balancing and extensive hyperparameter tuning, often failing to reproduce fine-grained details and exhibiting local artifacts under extreme pose or expression variations, which limit both naturalness and photorealism.


\paragrapht{Diffusion-based models.}
Diffusion models~\cite{ho2020ddpm, Song2020ddim, jascha2015, yang2019estgrad, yang2020sde, dhariwal2021diffusion, rombach2022high, dustin2023sdxl} have recently emerged as a powerful alternative to GANs, providing superior image quality and semantic controllability. Following this trend, various studies have adopted diffusion-based frameworks for face-swapping tasks.
DiffFace~\cite{kim2025diffface} first introduces an identity-conditional DDPM~\citep{ho2020ddpm} that incorporates multiple facial guidance signals, including semantic parsing and gaze direction. However, these cues are only applied during training and are not utilized at inference time, resulting in noticeable noise artifacts, particularly around the eyes.  
To alleviate this issue, subsequent approaches, such as FaceAdapter~\cite{han2024faceadapter}, DiffSwap~\cite{zhao2023diffswap}, and ReFace~\cite{yu2025reface}, reformulate the task as conditional inpainting, where the target facial region is masked and structural priors such as facial landmarks are used as conditioning signals.
Although this strategy helps prevent identity leakage from the target, masking out the entire face inevitably removes critical visual cues such as illumination, makeup, and accessories, making it difficult to preserve the target’s fine-grained attributes.
More recently, DreamID~\cite{ye2025dreamid} addresses this issue by constructing pseudo datasets using a GAN-based face-swapping model~\cite{rosberg2023facedancer} and training a diffusion network for identity transfer. While effective, it offers limited exploration into how to build high-quality pseudo triplets, particularly regarding attribute preservation. In contrast, our framework progressively mitigates these limitations through a teacher–student design. Crucially, we focus on improving the diffusion teacher itself so that it can generate high-fidelity, attribute preserving pseudo triplets, achieving significantly better performance than simply relying on an off-the-shelf face-swapping model.

\vspace{-5pt}
\section{Preliminaries}
\label{sec:preliminaries}

\paragraph{Rectified flow.}
Diffusion models~\cite{ho2020denoising, song2020score} learn to generate data by gradually denoising samples drawn from a Gaussian prior through a stochastic process. 
More recently, \textit{rectified flow}~\cite{liu2022flow} reformulated this process as a deterministic flow, simplifying sampling while maintaining distributional expressiveness.
Specifically, it defines a linear interpolation between a noise sample $\epsilon \sim \mathcal{N}(0,1)$ and a real sample $x_0 \sim p_0(x)$ at a timestep $t \in [0,1]$ as:
\begin{equation}
    \label{sec:diffusion_latent}
    z_t = (1 - t)\,x_0 + t\,\epsilon.
\end{equation}
The model learns to predict the velocity $v_t(x)$ that transports $x_0$ toward $\epsilon$ (or vice versa) along this linear path using a conditional flow-matching objective:
\begin{equation}
    \mathcal{L}_{\text{flow}} = 
    \mathbb{E}_{t,x_0\sim\,p_0(x)} \left[ 
    \| (\epsilon - x_0) - v_t(z_t) \|^2 
    \right].
\end{equation}
For the rectified flow models, intermediate prediction $\hat{x}_0$ is obtained by following equation:
\begin{equation}
    \hat{x}_0(z_t) = z_t - t\, v_t(z_t).
\end{equation}
Note that we omit the VAE encoder–decoder~\cite{rombach2021highresolution} for notational simplicity.

\paragrapht{Diffusion inversion.}
Inversion maps a real image back to its underlying Gaussian noise representation. When performed correctly, it yields an initial noise vector that can faithfully reconstruct the reference image, providing a reliable starting point for subsequent editing or manipulation. For rectified flow models, this can be accomplished similarly to DDIM inversion~\citep{Song2020ddim}. The latent trajectory is estimated by iteratively adding noise according to
\begin{equation}
z_{t+\Delta t} = z_t + \Delta t \cdot v_t(z_t),
\end{equation}
where $v_t(\cdot)$ denotes the learned velocity or score field.
By anchoring the generative path to real observations, inversion provides a crucial foundation for a wide range of image editing tasks~\citep{mokady2023null, hertz2022prompt, routsemantic, dengfireflow, xie2025dnaedit}.

\section{Methodology}
\label{sec:method}

\subsection{Overview}

Sec.~\ref{subsec:problem} formalizes the conventional conditional diffusion formulation for face swapping and defines the identity and attribute conditioning that our models follow. Sec.~\ref{subsec:teacher} presents our attribute-preserving diffusion teacher, trained with conditional deblurring. Sec.~\ref{subsec:pseudo} introduces attribute-aware inversion scheme to improve fine-grained attribute preservation. Sec.~\ref{subsec:student} then describes how a student diffusion model is trained with pseudo-label in a direct image-editing setting, leading to improved attribute fidelity while maintaining identity consistency.

\subsection{Problem formulation}
\label{subsec:problem}
Given a source and target image pair $\{I_{\text{src}}, I_{\text{tgt}}\}$, the face-swapping model aims to generate a swapped image $\hat{I}$ that reflects the identity of $I_{\text{src}}$ and the attributes of $I_{\text{tgt}}$, ensuring both identity fidelity and attribute consistency.


Recent diffusion-based face swapping methods~\citep{zhao2023diffswap, han2024faceadapter, ye2025dreamid} commonly follow a conditional diffusion framework in which identity and attribute information are injected into the model through separate conditioning pathways. To formalize this conventional design and establish a unified notation, we define two independent conditioning functions, each responsible for identity and attribute representation:
\begin{equation}
\mathbf{id}_{\text{src}} = \mathcal{F}_{\text{id}}(I_{\text{src}}), \qquad 
    \mathbf{att}_{\text{tgt}} = \mathcal{F}_{\text{att}}(I_{\text{tgt}}),
\end{equation}
where $\mathcal{F}_{\text{id}}(\cdot)$ extracts identity-related features (e.g., identity embeddings from a face recognition network), 
and $\mathcal{F}_{\text{att}}(\cdot)$ encodes attribute-related representations from the target (e.g., structural cues such as pose, expression, or lighting). 

During training, source-target image pair shares the same identity. Formally, the diffusion process for the face-swapping task is parameterized by a velocity field $v_t(\cdot)$ trained under a flow-matching objective that drives the model to reconstruct the target image conditioned on the source identity and target attributes:
\begin{equation}
    \mathcal{L}_{\text{flow}} 
    = \mathbb{E}_{t,\,I,\,\epsilon}\!
    \left[
        \|(\epsilon - I_\text{tgt}) - v_t(z_t,\, \mathbf{id}_{\text{src}},\, \mathbf{att}_{\text{tgt}})\|_2^2
    \right],
\end{equation}
where $\epsilon \!\sim\! \mathcal{N}(0, I)$ represents Gaussian noise, and $z_t$ is the interpolated latent between $I_\text{tgt}$ and $\epsilon$ at timestep $t$.

Following prior diffusion-based face swapping methods, we employ an identity loss to encourage the swapped output to match the source identity. The identity loss is defined as:
\begin{equation}
    \mathcal{L}_{\text{id}} =
1 - \cos\!\left(
\mathcal{F}_{\text{id}} \bigl(\hat{x_0}(z_t)\bigr),
\mathcal{F}_{\text{id}} \bigl(I_{\text{src}}\bigr)
\right),
\end{equation}
The overall training objective for the diffusion model combines the flow-matching loss with the identity loss:
\begin{equation}
    \mathcal{L}_{\text{total}} =
    \mathcal{L}_{\text{flow}}
+ \lambda_{\text{id}}\,\mathcal{L}_{\text{id}}.
\end{equation}

During inference, face swapping is achieved by using a source-target image pair with different identities.

\subsection{Teacher learning via conditional deblurring}
\label{subsec:teacher}

In our framework, the attributes include pose, facial expression, gaze direction, skin tone, makeup style, illumination, accessories, and background appearance. To ensure faithful preservation of these attributes, our framework introduces a carefully designed conditioning mechanism, enabling the teacher to preserve the aforementioned attributes while maintaining identity transferability.

\paragrapht{Conditional deblurring.}
Most diffusion-based face-swapping models~\citep{zhao2023diffswap,yu2025reface,han2024faceadapter} formulate the task as conditional \textit{inpainting}, where the facial region of the target image is masked and supervised to reconstruct target with source identity. Motivation for masking is to suppress identity information from the target image. Without masking, the model tends to reconstruct the target identity rather than learning to transfer the source identity.
While masking effectively prevents identity leakage, it also removes important attribute cues—such as lighting, skin tone, makeup, and accessories—that are crucial for faithful attribute preservation. 
As a result, inpainting-based diffusion models often fail to reproduce fine-grained target styles and yield inconsistent attribute appearance in the swapped face.

To overcome this limitation, we reformulate the training objective as a conditional \textit{deblurring} task. 
Instead of masking the entire facial region, we replace it with a blurred version of the target image, which removes high-frequency identity details while retaining low-frequency but informative attribute cues, including pose, lighting, and expression. 
This reformulation allows the model to exploit the target’s contextual attribute information more effectively during training, leading to superior preservation of lighting, skin tone, and structural attributes while maintaining strong identity transferability, as shown in Fig.~\ref{fig:condition_comparison}

Specifically,
we downsample the target image to a resolution of $8\times8$ and then upsample it back to the original size, effectively eliminating fine-grained identity details while preserving low-frequency appearance such as color tone and lighting. A facial mask extracted from a face parsing model~\citep{voo2022delving} is then used to apply the blurring only to the facial region, ensuring that background and contextual information remain unaffected.



\paragrapht{Enriching semantic conditions.} To maintain structural consistency and realism, we enrich the semantic conditions by incorporating multi-level structural cues beyond standard 3DMM~\citep{blanz2023morphable_3dmm} landmarks. While gaze is a critical factor, existing diffusion-based models often struggle to preserve gaze alignment or introduce artifacts when using gaze loss as sampling guidance. To address these limitations, we overlay eye landmarks derived from a gaze estimator~\cite{ablavatski2020real} and glass segmentation masks obtained via face parsing~\cite{yu2018bisenet} onto the blurred target condition. This integrated approach provides the diffusion model with both coarse attribute context and fine structural details, enabling the faithful preservation of pose, gaze, and accessories during face swapping.

\subsection{Pseudo-label generation with inversion}
\label{subsec:pseudo}
Although the proposed deblurring strategy and semantic condition used in training the teacher model effectively improve attribute preservation, there remains room to enhance further the reconstruction of fine-grained details such as makeup and accessories. This is because the model is required to implicitly infer high-frequency details that are not explicitly present in the blurred inputs.

\paragrapht{Attribute-aware inversion.}
To address this issue, we draw inspiration from inversion-based editing techniques~\citep{mokady2023null, hertz2022prompt, routsemantic, dengfireflow, xie2025dnaedit}, which enable more precise control over fine-grained attributes by aligning the generation process with latent representations.
Recent studies have shown that noise obtained through diffusion inversion deviates from the ideal Gaussian prior~\citep{staniszewski2024there, ahn2024noise} and retains residual semantic information from the input, such as image structure or prompt-related signals~\citep{staniszewski2024there, li2024source}. While prior work seeks to suppress this residual information due to its negative impact on editability~\citep{staniszewski2024there}, we instead exploit it.
In face swapping, where the goal is to modify identity while preserving target-specific attributes, retaining such information in the noise is beneficial rather than detrimental.

Motivated by this insight, we propose \textit{attribute-aware inversion} which inverts the target image under attribute-only conditioning, yielding noise representations that are enriched with target attributes but free from target identity bias. Using these attribute-aware noise at inference time enables the model to better preserve fine-grained appearance characteristics that are often lost due to blurring, while maintaining strong identity transferability.

\begin{figure}[t]
    \centering
    \includegraphics[width=\linewidth]{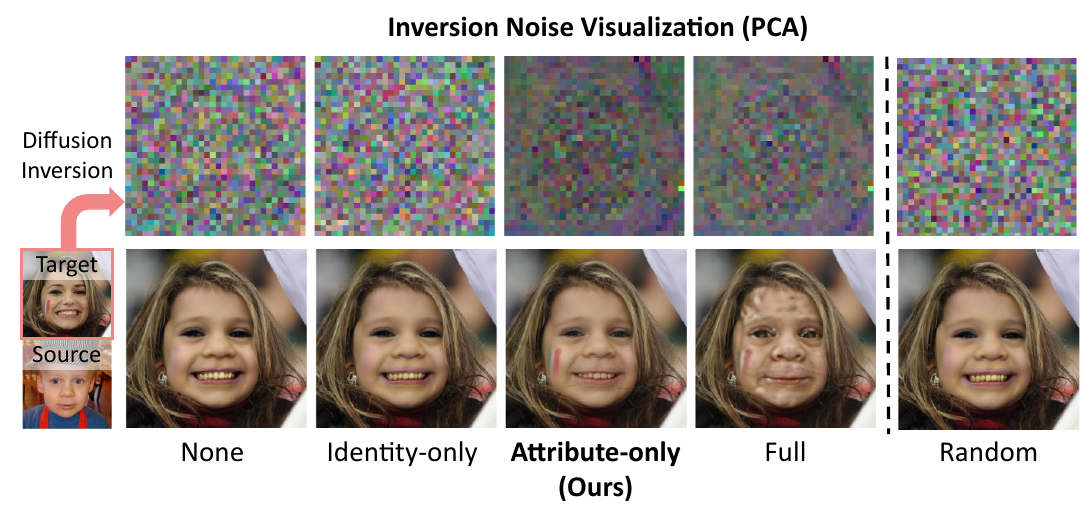} 
    \vspace{-20pt}
    \captionof{figure}{
        \textbf{Comparison of conditioning configuration for inversion.} (Top) PCA visualization of target-inverted noise and random Gaussian noise. When attribute-only conditioning is used, inverted noise encodes more semantic information compared to the others. (Bottom) Results of face swapping when each inverted noise is used. Using attribute-only conditioned noise yields the most makeup preserved results without introducing artifacts.  }
    \label{fig:inversion_comparison}
\vspace{-10pt}
\end{figure}

To validate our choice, we conduct an empirical analysis comparing four conditioning configurations that can be used in inversion, formally defined as
\[
(\mathcal{F}_{\text{id}}(I), \mathcal{F}_{\text{att}}(I)),
(\varnothing, \mathcal{F}_{\text{att}}(I)),
(\mathcal{F}_{\text{id}}(I), \varnothing),
(\varnothing,\varnothing)
\]
where each element corresponds to full conditioning, attribute-only conditioning, identity-only conditioning, and non-conditioning.

We first analyze how each configuration affects the inverted noise by performing PCA on the resulting noise representations. The non-conditioned and identity-only configurations produce noise distributions with little semantic structure, similar to random noise. In contrast, configurations that incorporate attribute conditioning yield noise patterns that exhibit clear facial semantics, indicating that attribute information is effectively retained during inversion.

We additionally compare the swapped outputs obtained from each inversion configuration. Both attribute-only and full conditioning preserve fine-grained attributes such as makeup more reliably than the other variants. However, full conditioning often produces artifacts in the swapped images. We attribute this to residual identity information embedded in the full-conditioned inverted noise, which restricts editability and interferes with identity replacement. In contrast, attribute-only conditioning $(\varnothing, \mathcal{F}_{\text{att}}(I))$ avoids identity-related biases while still embedding meaningful attribute cues. This allows the model to preserve target attributes better during swapping, without introducing artifacts. Based on these observations, we adopt attribute-only conditioning as our inversion strategy.

\subsection{Student learning with pseudo-triplet}
\label{subsec:student}

We employ the teacher model to synthesize a pseudo-label with the attribute-aware inversion scheme.  Specifically, by performing face swapping on a target image $I_{\text{tgt}}^{A}$ of identity $A$ with another subject $B$ via the teacher model, we obtain a pseudo-label $\hat{I}_{\text{tgt}}^{A \rightarrow B}$ that is used to construct the pseudo-triplet $(I_{\text{src}}^{A}, \hat{I}_{\text{tgt}}^{A \rightarrow B}, I_{\text{tgt}}^{A})$.  The triplet is then used to train the student under an explicit editing objective. Specifically, the student takes an identity feature from source image $I_{\text{src}}^{A}$ and an attribute feature from pseudo-label $I_{\text{tgt}}^{A \rightarrow B}$ as input and learns to reconstruct the original target $I_{\text{tgt}}^{A}$.

By designing the student model to accept pseudo-labels in their raw image format as attribute conditioning inputs, we achieve two significant benefits. First, during the training phase, the model leverages these high-fidelity, unmasked inputs instead of the degraded images used in conventional approaches. This allows the student model to learn superior attribute-preservation more effectively. Second, this design removes the need for auxiliary networks or complex preprocessing pipelines for attribute conditioning, thereby improving practicality in real-world deployment.

\section{Experiments}


\begin{figure*}[t] 
    \centering
    \includegraphics[width=\textwidth]{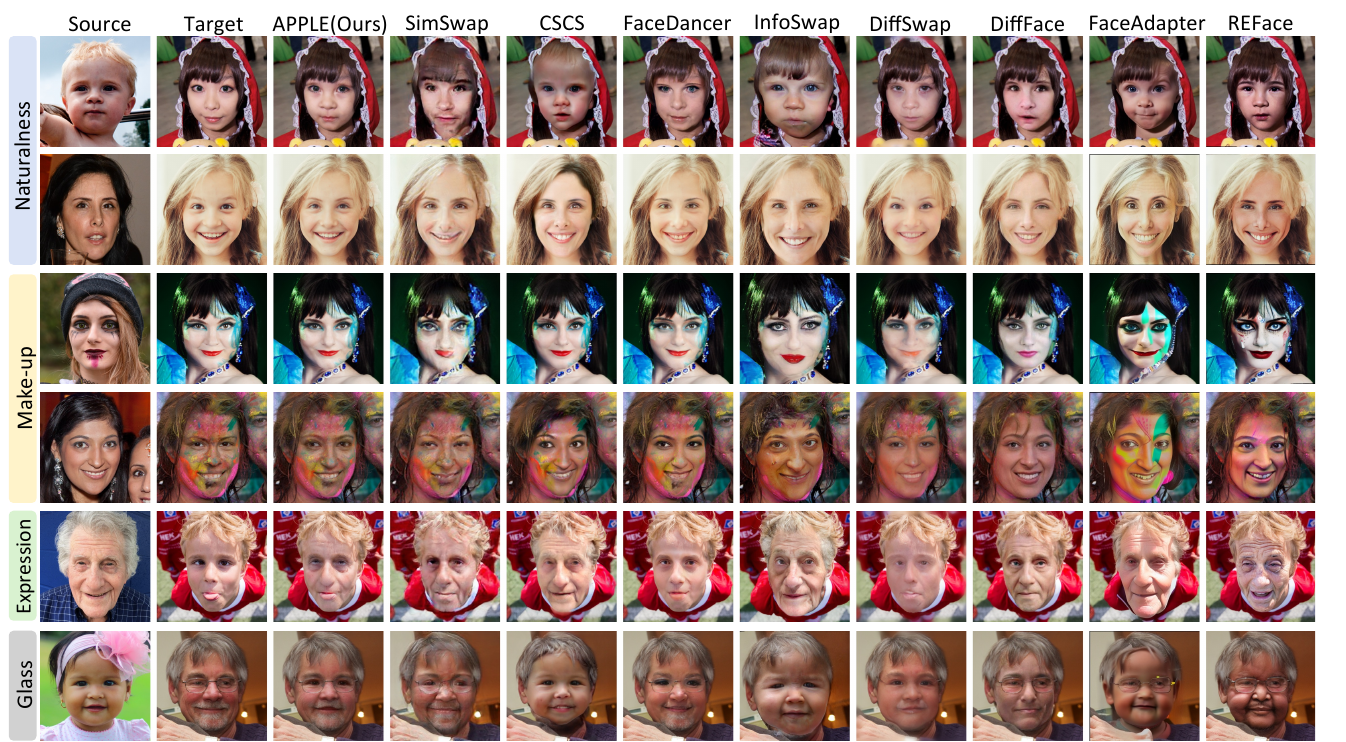}
    \caption{\textbf{Qualitative results on FFHQ.} 
    Compared to existing baselines, \ourframework effectively preserves the target image’s attributes while faithfully transferring the source identity.}
    \label{fig:main_qual}
    \vspace{-10pt}
\end{figure*}

\subsection{Experimental setups}

\paragrapht{Implementation details.}
We use FLUX.1-Krea [dev]~\cite{flux2024} as the base diffusion backbone for all experiments, employing PulID~\cite{guo2024pulid} as the identity encoder and OminiControl~\cite{tan2025ominicontrol} as the attribute conditioning branch with LoRA rank of 64. The model is trained on the VGGFace2-HQ dataset~\cite{chen2023simswap++} which is filtered with AES~\cite{schuhmann2022aestheticpredictor} to ensure high-quality faces, with threshold 5.1. Following REFace~\cite{yu2025reface}, the source image is masked before being fed into the identity encoder. 
Teacher model is trained for 15K iterations without the identity loss and an additional 50K iterations with it enabled. Student model is trained for 15K iterations, resuming from teacher. Experiments are conducted on four NVIDIA A6000 GPUs with a batch size of 1 per GPU, gradient accumulation of 4, resulting in effective batch size 16.

\paragrapht{Evaluation protocol.}
Following previous studies~\cite{huang2024identity, ye2025dreamid}, we select 1,000 source faces and 1,000 target faces from the FFHQ dataset~\cite{karras2019style}, and generated 1,000 corresponding face-swapped results. 
To assess image fidelity, we compute Frechet inception distance(FID) between the face-swapped images and the real FFHQ images. 
For pose and expression consistency, we employ HopeNet~\cite{doosti2020hope} and Deep3DFaceRecon~\cite{deng2019accurate}, respectively, and measure the L2 distance between each target image and its corresponding swapped image. 
To evaluate identity preservation, ArcFace~\cite{deng2019arcface} is to extract identity embeddings, and the cosine distance is calculated between the embeddings of the swapped and source faces. 
To compute ID Retrieval,  we find the most similar source faces based on cosine similarity, and report Top-1 and Top-5 accuracy. 
All experiments, including training and inference, are performed on images of resolution $512 \times 512$.

\begin{table}[t]
\centering
\caption{\textbf{Ablation of teacher model.}
Applying the proposed components sequentially leads to gradual performance improvements. In particular, the attribute-aware inversion module substantially improves FID, pose, and expression metrics while maintaining identity similarity at a comparable level.}

\vspace{-5pt}

\resizebox{\linewidth}{!}{
    \begin{tabular}{l|ccccc}
    \toprule

    \multirow{2}{*}{\textbf{Model}} 
    & \multirow{2}{*}{\textbf{FID}$\downarrow$} 
    & \multirow{2}{*}{\textbf{ID Sim.}$\uparrow$} 
    & \textbf{ID Retrieval}$\uparrow$ 
    & \multirow{2}{*}{\textbf{Pose}$\downarrow$} 
    & \multirow{2}{*}{\textbf{Expr.}$\downarrow$} \\

    &  &  & \textbf{Top-1 / Top-5} &  &  \\
    \midrule
    Inpainting (Baseline) & 11.00  & 0.54 & 92.80 / 96.90 & 3.37 & 1.01   \\
    Deblurring & 4.20  & 0.53 & 89.50 / 96.20 & 2.58 & 0.79  \\
    Deblurring + Inv. & \textbf{3.68} & 0.54 & 90.40 / 96.70 & \textbf{2.07} & \textbf{0.70} \\
    \bottomrule
    \end{tabular}
}

\label{tab:ablation_teacher}


\end{table}

\begin{table}[t]
\centering

\caption{\textbf{Ablation of conditioning configurations for inversion.} 
For inversion, using attribute-only condition improves FID, pose and expression effectively while not sacrificing identity transferability. Applying the other conditions does not robustly improve FID, pose or expression.
}

\vspace{-5pt}

\resizebox{\linewidth}{!}{
    \begin{tabular}{l|ccccc}
    \toprule

    \multirow{2}{*}{\textbf{Condition config.}} 
    & \multirow{2}{*}{\textbf{FID}$\downarrow$} 
    & \multirow{2}{*}{\textbf{ID Sim.}$\uparrow$} 
    & \textbf{ID Retrieval}$\uparrow$ 
    & \multirow{2}{*}{\textbf{Pose}$\downarrow$} 
    & \multirow{2}{*}{\textbf{Expr.}$\downarrow$} \\

    &  &  & \textbf{Top-1 / Top-5} &  &  \\
    
    \midrule
    Baseline & 4.20  & 0.53 & 89.50 / 96.20 & 2.58 & 0.79  \\
    \midrule
    + None & 6.20  & 0.52 & 87.90 / 95.80 & \textbf{2.03} & 0.74  \\
    + Identity-only & 10.02 & 0.53 & 90.50 / 95.90 & 2.57 & 0.83 \\
    + Attribute-only & \textbf{3.68} & 0.54 & 90.40 / 96.70 & 2.07 & \textbf{0.70} \\
    + Full & 10.51 & 0.53 & 89.10 / 93.80 & 3.13 & 0.99 \\
    \bottomrule
    \end{tabular}
}

\label{tab:ablation_inversion}
\end{table}

\begin{table}[!t]
\centering

\caption{\textbf{Ablation of pseudo-label quality.}
We compare the performance of student models trained on pseudo datasets generated by FaceDancer~\citep{rosberg2023facedancer} and by our teacher model. The model trained on our pseudo-triplet achieves superior attribute-preserving results indicated by lower pose and expression consistency, while maintaining comparable identity similarity. Results indicate that our teacher model serves as a more effective teacher.
}

\vspace{-5pt}

\resizebox{\linewidth}{!}{
    \begin{tabular}{lccccc}
    \toprule
    \textbf{Teacher model} & \textbf{FID}$\downarrow$ & \textbf{ID Sim.}$\uparrow$ & \textbf{ID Retrieval}$\uparrow$ & \textbf{Pose}$\downarrow$ & \textbf{Expr.}$\downarrow$ \\
    \midrule
    FaceDancer \citep{rosberg2023facedancer} & 2.47  & 0.53 & 87.50 / 95.60 & 2.07 & 0.65  \\
    \textbf{\ourframework (Teacher)} & \textbf{1.98} & 0.53 & 88.60 / 95.80 & \textbf{1.77} & \textbf{0.62}  \\
    \bottomrule
    \end{tabular}
}


\label{tab:ablation_data}
\end{table}

\begin{figure}[t] 
    \centering
    \includegraphics[width=\columnwidth]{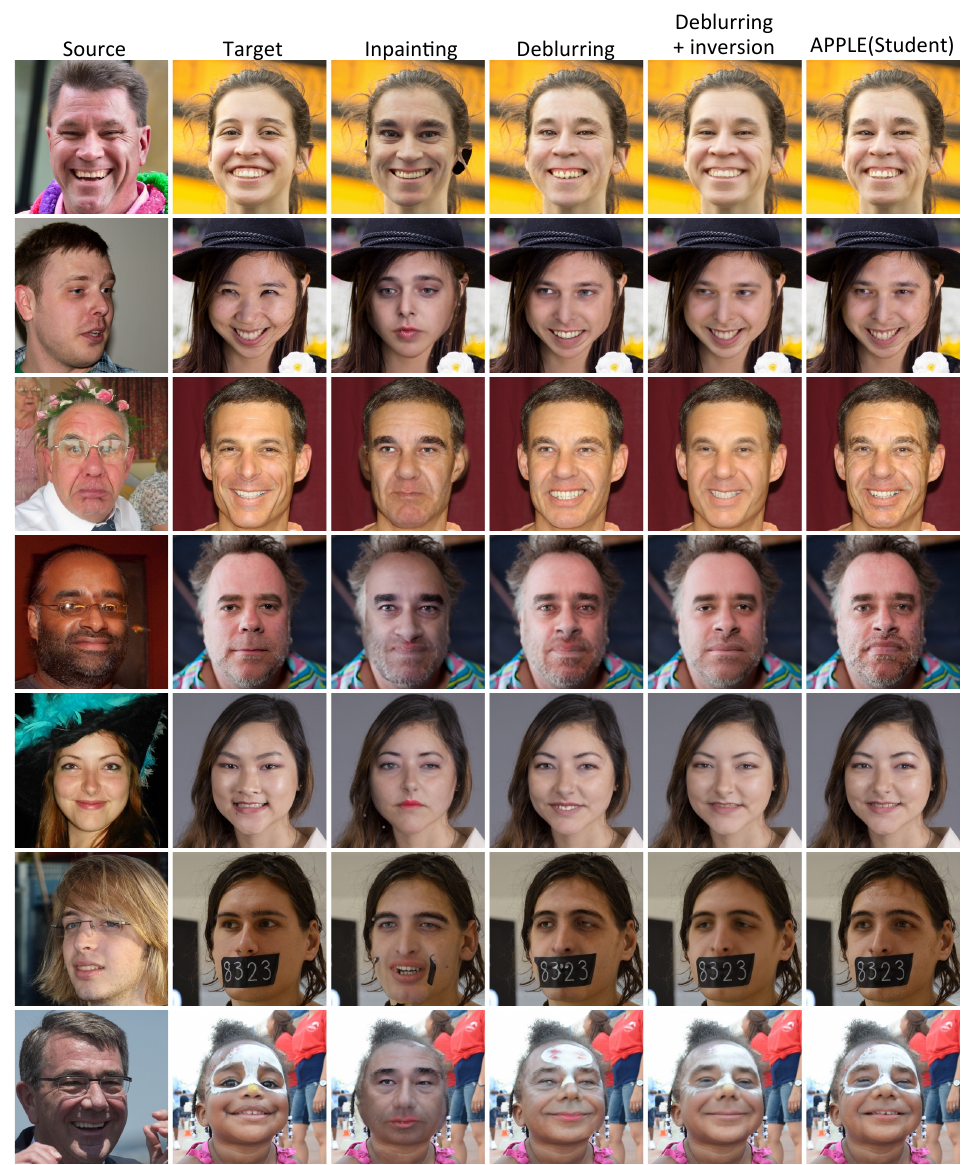}
    \caption{\textbf{Qualitative ablation results of proposed method.}
     The inpainting method fails to preserve the target image’s attributes. Applying deblurring (4th column) and attribute-aware inversion (5th column) progressively better maintain the target attributes. Overall quality, including attribute-preserving is maximized on student (6th column).
    } 
    \label{fig:ablation_teacher_qual}
\end{figure}

\subsection{Comparison with other models}
We compare our method both diffusion-based and GAN-based approaches. Across all experiments, our model delivers substantially stronger attribute preservation while maintaining identity similarity competitive with the best existing methods. It is worth noting that several baselines, including DiffSwap~\citep{zhao2023diffswap}, DiffFace~\citep{kim2025diffface}, and E4S~\citep{liu2023fine_e4s}, are trained directly on FFHQ and therefore benefit from favorable domain alignment with the evaluation benchmark. 

\begin{table}[t]
\centering

\caption{\textbf{Quantitative results on FFHQ.}
While maintaining identity transferability, \ourframework achieves superior attribute preserving performance, which is evidenced by lowest pose and expression error. 
}
\vspace{-5pt}

\resizebox{\linewidth}{!}{
    \begin{tabular}{lccccc}
    \toprule

    \multirow{2}{*}{\textbf{Model}} 
    & \multirow{2}{*}{\textbf{FID}$\downarrow$} 
    & \multirow{2}{*}{\textbf{ID Sim.}$\uparrow$} 
    & \textbf{ID Retrieval}$\uparrow$ 
    & \multirow{2}{*}{\textbf{Pose}$\downarrow$} 
    & \multirow{2}{*}{\textbf{Expr.}$\downarrow$} \\

    &  &  & \textbf{Top-1 / Top-5} &  &  \\
    \midrule
    SimSwap~\cite{chen2020simswap} & 18.54 & 0.55 & 94.10 / 99.00  & 3.11 &  1.73   \\
    CSCS~\cite{huang2024identity_cscs} & 11.00 & 0.65 & 99.00 / 99.50 & 3.64 &  1.44 \\
    MegaFS~\cite{zhu2021one_megafs} & 12.83 & 0.49 & 79.6 / 86.3 & 4.40 &  1.11 \\
    FaceDancer~\cite{rosberg2023facedancer} & 3.80 & 0.51 & 89.70 / 96.50 & 2.23 & 0.74 \\
    HiFiFace~\cite{wang2021hififace} & 11.81 & 0.50 & 85.40 / 93.40 & 3.20 & 1.34 \\
    InfoSwap~\cite{gao2021information_infoswap} & 5.00 & 0.54 & 91.40 / 97.50 & 4.33 & 1.40 \\
    E4S~\cite{liu2023fine_e4s} & 12.13 & 0.49 & 78.30 / 87.80 & 4.39 & 1.29 \\
    \midrule
    DiffSwap~\cite{zhao2023diffswap} & 6.84 & 0.34 & 41.92 / 63.09 & 2.63 & 1.20 \\
    DiffFace~\cite{kim2025diffface} & 8.59 & 0.54 & 90.70 / 95.90 & 3.67 & 1.24 \\
    FaceAdapter~\cite{han2024faceadapter} & 13.03 & 0.52 & 87.00 / 93.20 & 5.12 & 1.38 \\
    REFace~\cite{yu2025reface} & 7.22 & 0.60  &  97.60 / 99.40 & 3.67 & 1.08 \\
    \midrule
    \textbf{\ourframework (Teacher)}  & 3.68 & 0.54 & 90.40 / 96.70 & 2.07 & 0.70  \\
    \textbf{\ourframework (Student)} & \textbf{2.18} & 0.54 & 90.50 / 97.00 & \textbf{1.85} & \textbf{0.64} \\
    \bottomrule
    \end{tabular}
}

\label{tab:ffhq_comparison}
\vspace{-10pt}
\end{table}

\paragrapht{Qualitative comparison.}
We present qualitative results in Fig.~\ref{fig:main_qual}. 
Our model maintains subtle cues such as tongue-out \textit{expressions}, fine-grained \textit{makeup} details, and \textit{accessory} (e.g. glass), while still performing reliable identity transfer. Competing diffusion-based methods often fail in these scenarios by either erasing attributes or inventing new ones. GAN-based models  preserve attributes better than existing diffusion-based models because their generators observe the full target image during self-supervised training. However, their outputs frequently contain artifacts, color inconsistencies, and unnatural textures due to the inherent limitations of adversarial training. In contrast, our method produces clean and realistic results while retaining the attribute fidelity even better than GAN based models.

\paragrapht{Quantitative comparison.}
The quantitative results reported in Tab.~\ref{tab:ffhq_comparison} follow the similar trend observed in the qualitative results. Our method achieves the lowest FID among all baselines, indicating cleaner and more realistic synthesis, and shows clear improvements in pose and expression consistency while maintaining competitive identity similarity. Although CSCS~\citep{huang2024identity_cscs} and REFace~\citep{yu2025reface} report notably high identity similarity, this comes with a substantial drop in attribute preservation, evidenced by the copy-paste–like artifacts visible in their qualitative outputs in Fig.~\ref{fig:main_qual}. Identity transfer and attribute preservation inherently trade off against each other, so a well-designed method should balance the two. However, they are strongly biased toward identity matching, leading to poor attribute fidelity, which is an undesirable behavior for face swapping. In contrast, \ourframework achieves a more balanced trade-off, delivering reliable identity transfer while consistently preserving target attributes across diverse scenarios.

\vspace{-5pt}
\subsection{Ablation studies}


In this section, we analyze the effectiveness of our teacher-student framework through component-wise ablations.



\paragrapht{Inpainting versus deblurring.} We compare the conventional inpainting with our deblurring formulation. The qualitative and quantitative results are presented in Fig.~\ref{fig:ablation_teacher_qual} and Tab.~\ref{tab:ablation_inversion}. Qualitative results show that transition from inpainting to deblurring substantially improves attribute preservation. Deblurring produces noticeably more consistent skin tone, lighting, pose, and expression, since the model is no longer required to synthesize these cues. These qualitative gains are reflected in quantitative metrics, which show huge improvements in pose and expression consistency. 

\paragrapht{Effect of attribute-aware inversion.} We additionally study the impact of attribute-aware inversion qualitatively and quantitatively, presenting results presented in Fig.~\ref{fig:ablation_teacher_qual} and Tab.~\ref{tab:ablation_inversion}. When proposed attribute-aware inversion is used, the model better reconstructs fine-grained attributes such as makeup, jewelry, and accessories, which are difficult to recover from blurred observations alone. Quantitative results confirm that attribute-aware inversion further improves pose and expression without sacrificing identity transferability.

\begin{figure}[t] 
    \centering
    \includegraphics[width=\columnwidth]{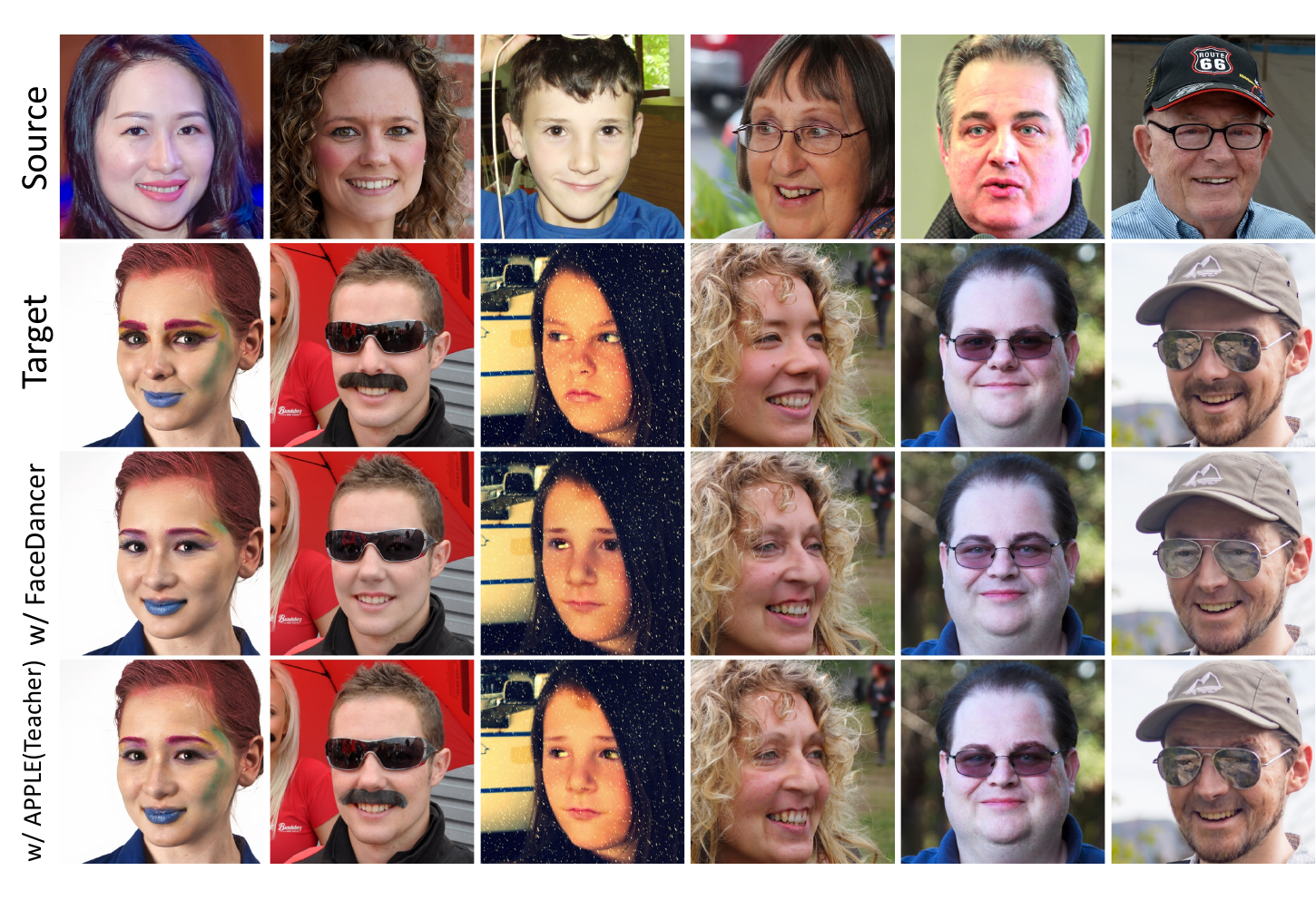}
    \caption{\textbf{Qualitative ablation results of pseudo-label quality.} We compare two student models trained on pseudo-triplet generated by FaceDancer~\cite{rosberg2023facedancer} and APPLE (Teacher). Results demonstrate that student trained by APPLE (Teacher) more faithfully preserves the target attributes.}
    \label{fig:ablation_data}
\end{figure}

\paragrapht{Various inversion strategies.} 
We conduct a quantitative ablation on the conditioning configurations used during the inversion process. Specifically, we evaluate the result of teacher model using four types of inversion conditions and report the results in Tab.~\ref{tab:ablation_inversion}. The results show that using attribute-only condition $(\varnothing, \mathcal{F}_{\text{att}}(I))$ yields the robust performance, whereas the other variants introduce noticeable side effects such as increased FID, pose or expression.

\paragrapht{Quality of pseudo-label.}
We further evaluate the quality of our pseudo-triplet by comparing student models trained on different pseudo-triplets. Since DreamID~\citep{ye2025dreamid} adopts FaceDancer~\citep{rosberg2023facedancer} as a teacher due to its strong attribute preservation, we compare a student trained on FaceDancer-generated pseudo-triplet with one trained on pseudo-triplet produced by our teacher. Both 50K pseudo-triplets are generated from VGGFace2-HQ~\citep{chen2023simswap++}, and all models are evaluated at 5K training steps. While the FaceDancer-based student achieves slightly higher identity similarity, the student trained on our pseudo-triplet performs better on pose, expression, and fine-grained attribute metrics. Qualitative results in Fig.~\ref{fig:ablation_data} further show that our pseudo-triplet leads to more accurate preserving of attributes such as makeup, gaze, accessories etc. This demonstrates that pseudo-triplet generated by our teacher provides stronger supervision for training high-quality face-swapping models.

\section{Conclusion}
\label{sec:conclusion}

In this work, we introduced \textbf{\ourframework}, a diffusion-based teacher–student framework that tackles the challenge of attribute preservation in face swapping. We designed a training and inference strategy that produces attribute-preserving pseudo-labels, enabling the student to learn in a direct image-editing setting with a clean attribute condition image. This allows the student to retain fine-grained details more faithfully than its teacher and existing diffusion- and GAN-based methods, while maintaining competitive identity transferability. Our results demonstrated that the proposed teacher–student design provides an effective path toward high-fidelity, attribute-preserving diffusion face swapping without sacrificing identity transferability.

\section*{Acknowledgements}
This research was supported by Institute of Information \& communications Technology Planning \&
Evaluation (IITP) grant funded by the Korea government (MSIT) (RS-2019-II190075, RS-2024-00509279, RS-2025-II212068, RS-2023-00227592, RS-2025-02214479, RS-2024-00457882, RS-2025-25441838, RS-2025-25441838, RS-2025-02214479, RS-2025-02217259) and the Culture, Sports, and Tourism R\&D Program through the Korea Creative Content Agency grant funded by the Ministry of Culture, Sports and Tourism (RS-2024-00345025, RS-2024-00333068, RS-2023-00222280, RS-2023-00266509), and National Research Foundation of Korea (RS-2024-00346597).

{
    \small
    \bibliographystyle{ieeenat_fullname}
    \bibliography{main}
}


\clearpage
\appendix

\twocolumn
\noptcrule  
\setcounter{page}{1}

\twocolumn[
    
{
    
    \begin{center}
        \textbf{\Large Attribute-Preserving Pseudo-Labeling for Diffusion-Based Face Swapping}
        \vspace{1em} \\
       {\large Supplementary Material}
    \end{center}

    \doparttoc
    \faketableofcontents
    \part{}
    
    \vspace{-100pt}
    
    \mtcsettitle{parttoc}{}
    \part{}
    \noindent\rule{\linewidth}{0.4pt}\par   
    \parttoc    
    \noindent\rule{\linewidth}{0.4pt}\par   
    
\vspace{10pt}

\begin{center}
    \includegraphics[width=\textwidth]{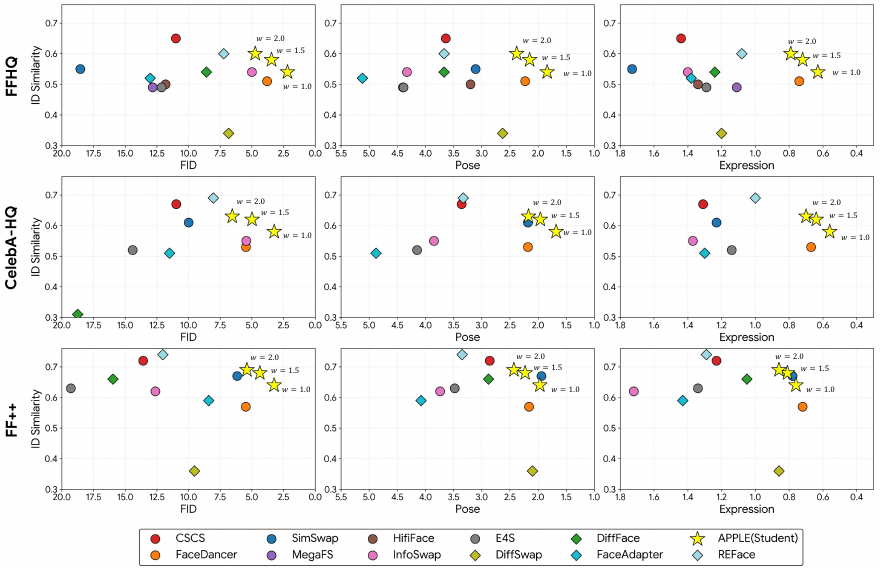}
    \captionof{figure}{\textbf{Pareto frontier visualizations across datasets.} We visualize ID Sim. vs. FID, Pose, and Expression reported in \cref{tab:ffhq_comparison,tab:guidance_scale_ablation,tab:celebahq_comparison,tab:faceforensicspp_comparison}. Ideal face-swapping models should occupy the \textbf{upper-right region}---high identity similarity with low FID and pose/expression error. APPLE(Student) \emph{consistently} lies on or near the Pareto frontier across all datasets, demonstrating \textbf{robust attribute preservation}. For APPLE, we present results under varying guidance scale $w$. Note that some baselines (REFace, CSCS, InfoSwap) are trained on CelebA-HQ.}
    \label{fig:eval_additional_dset}
\end{center}
}
]


\clearpage

\section{Experimental Details}
\subsection{Additional implementation details}
We use FLUX.1-Krea [dev]~\citep{flux2024} as the base diffusion backbone for all experiments, employing PulID~\citep{guo2024pulid} as the identity encoder and OminiControl~\citep{tan2025ominicontrol} as the attribute conditioning branch with a LoRA~\citep{hu2022lora} rank of 64. The model is trained on the VGGFace2-HQ~\citep{chen2023simswap++} dataset and evaluated on FFHQ~\citep{karras2019style}. Models are trained on images filtered by an AES~\citep{schuhmann2022aestheticpredictor} $\geq$ 5.1 threshold to ensure high-quality faces. Following REFace~\citep{yu2025reface}, the source image is masked before being fed into the identity encoder. The overall objective combines identity, flow-matching losses. Identity losses are applied to late 35\% of the diffusion timesteps for teacher model and late 50\% for student model. Teacher model is trained for 15K iterations without the identity loss and for 50K iterations with it enabled. Student model is resumed from teacher model and trained for 15K steps with pseudo-triplet generated by teacher model. Experiments are mainly conducted on four NVIDIA A6000 GPUs with a batch size of 1 per GPU, gradient accumulation of 4, resulting in an effective batch size of 16. AdamW is used as optimizer with learning rate 1e-4 and weight decay of 1e-3 is used. Inference and inversion step of diffusion model is fixed to 28. Training and inference are conducted at a resolution of 512x512.

\section{Additional Experiments}

\subsection{Evaluation on additional datasets with diverse guidance scales}
\paragrapht{Guidance scale ablation on FFHQ.}
Since our model $v_t(\cdot)$ accepts source identity feature $\mathbf{id}_{\mathrm{src}}$ and target attribute feature $\mathbf{att}_{\mathrm{tgt}}$ as conditions, classifier-free guidance (CFG) with scale $w$ can be applied to the identity condition to strengthen identity transferability:

\begin{equation}
\begin{aligned}
\tilde{v}_t =\ & v_t(z_t, \varnothing, \mathbf{att}_{\mathrm{tgt}}) \\
& + w \Bigl(
v_t(z_t, \mathbf{id}_{\mathrm{src}}, \mathbf{att}_{\mathrm{tgt}})
- v_t(z_t, \varnothing, \mathbf{att}_{\mathrm{tgt}})
\Bigr).
\end{aligned}
\end{equation}

\begin{figure*}[t]
    \centering
    \includegraphics[width=1.0\linewidth]{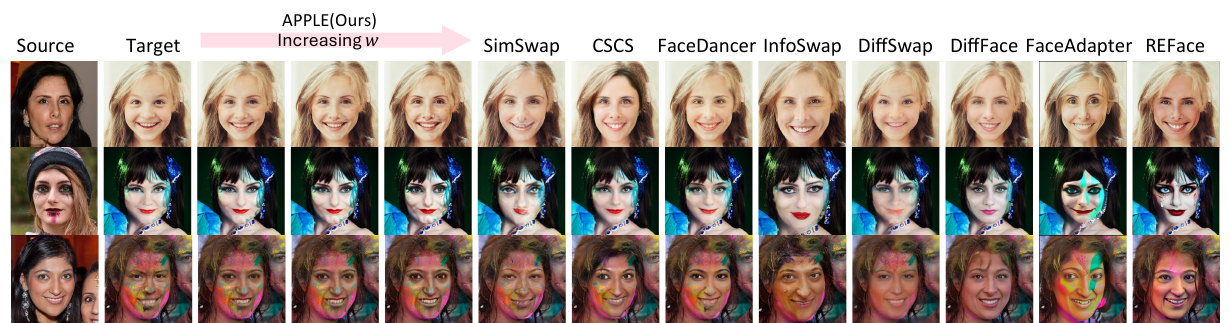}
    \vspace{-20pt}
    \caption{\textbf{Ablation study of guidance scale $w$ on FFHQ.} Guidance scale $w$ enables inference-time control of the identity-attribute trade-off. From our default $w=1.0$, increasing $w$ boosts identity transferability at the cost of moderate attribute preservation degradation.}
    \label{fig:guidance_ablation}
\end{figure*}

We report results of FFHQ across varying guidance scales $w$ in Tab.~\ref{tab:guidance_scale_ablation} and Fig.~\ref{fig:guidance_ablation}. Starting from the default setting $w=1.0$, increasing $w$ improves identity transferability at the cost of moderate degradation in attribute preservation, demonstrating inference-time control over the trade-off. Thus, user can flexibly adjust $w$ to prioritize either identity transfer or attribute preservation based on specific application needs.

Note that while other diffusion-based baselines (DiffFace, FaceAdapter, REFace) are evaluated using their respective tuned hyperparameter $w$, \textbf{we report our main results without hyperparameter tuning ($w=1.0$)}. 

\begin{table}[t]
\centering

\caption{\textbf{Ablation study of guidance scale $w$ on FFHQ.} Unless otherwise specified, all results in the main paper use $w=1.0$ by default. Increasing the guidance scale enhances identity similarity and retrieval performance, albeit with a moderate trade-off in attribute preservation metrics, such as pose and expression.}

\vspace{-5pt}

\resizebox{\linewidth}{!}{
    \begin{tabular}{lccccc}
    \toprule
    \multirow{2}{*}{\textbf{Model}}
    & \multirow{2}{*}{\textbf{FID}$\downarrow$}
    & \multirow{2}{*}{\textbf{ID Sim.}$\uparrow$}
    & \textbf{ID Retrieval}$\uparrow$
    & \multirow{2}{*}{\textbf{Pose}$\downarrow$}
    & \multirow{2}{*}{\textbf{Expr.}$\downarrow$} \\
    & & & \textbf{Top-1 / Top-5} & & \\
    \midrule
    \textbf{\ourframework (Student)}, $w$ = 1.0 & 2.19 & 0.54 & 90.40 / 96.50 & 1.84 & 0.63 \\
    \textbf{\ourframework (Student)}, $w$ = 1.5 & 3.45 & 0.58 & 94.60 / 98.30 & 2.15 & 0.72 \\
    \textbf{\ourframework (Student)}, $w$ = 2.0 & 4.74 & 0.60 & 96.00 / 98.40 & 2.38 & 0.79 \\
    \bottomrule
    \end{tabular}
}

\label{tab:guidance_scale_ablation}
\end{table}

\begin{table}[t]
\centering

\caption{\textbf{Quantitative results on CelebA-HQ.}
Note that some baselines (REFace, CSCS, InfoSwap) are trained on CelebA-HQ, while APPLE is trained solely on VGGFace2-HQ.
}
\vspace{-5pt}

\resizebox{\linewidth}{!}{
    \begin{tabular}{lccccc}
    \toprule
    \multirow{2}{*}{\textbf{Model}}
    & \multirow{2}{*}{\textbf{FID}$\downarrow$}
    & \multirow{2}{*}{\textbf{ID Sim.}$\uparrow$}
    & \textbf{ID Retrieval}$\uparrow$
    & \multirow{2}{*}{\textbf{Pose}$\downarrow$}
    & \multirow{2}{*}{\textbf{Expr.}$\downarrow$} \\
    & & & \textbf{Top-1 / Top-5} & & \\
    \midrule
    SimSwap & 9.99 & 0.61 & 97.70 / 99.00 & 2.18 & 1.23 \\
    CSCS & 10.97 & 0.67 & 97.70 / 99.50 & 3.36 & 1.31 \\
    FaceDancer & 5.47 & 0.53 & 92.60 / 98.50 & 2.18 & 0.67 \\
    InfoSwap & 5.44 & 0.55 & 93.10 / 97.50 & 3.85 & 1.37 \\
    E4S & 14.41 & 0.52 & 83.50 / 92.30 & 4.15 & 1.14 \\
    \midrule
    DiffSwap & 20.08 & 0.23 & 18.94 / 35.07 & 7.54 & 2.08 \\
    DiffFace & 18.75 & 0.31 & 46.30 / 47.30 & 10.84 & 1.88 \\
    FaceAdapter & 11.50 & 0.51 & 85.10 / 90.80 & 4.88 & 1.30 \\
    REFace & 8.04 & 0.69 & 99.10 / 99.90 & 3.33 & 1.00 \\
    \midrule
    \textbf{\ourframework (Student)}, $w$ = 1.0 & 3.24 & 0.58 & 95.80 / 98.60 & 1.68 & 0.56 \\
    \textbf{\ourframework (Student)}, $w$ = 1.5 & 4.99 & 0.62 & 97.00 / 99.30 & 1.96 & 0.64 \\
    \textbf{\ourframework (Student)}, $w$ = 2.0 & 6.55 & 0.63 & 97.30 / 99.50 & 2.17 & 0.70 \\
    \bottomrule
    \end{tabular}
}

\label{tab:celebahq_comparison}
\end{table}

\begin{table}[!t]
\centering

\caption{\textbf{Quantitative results on FaceForensics++.}}
\vspace{-5pt}

\resizebox{\linewidth}{!}{
    \begin{tabular}{lccccc}
    \toprule
    \multirow{2}{*}{\textbf{Model}}
    & \multirow{2}{*}{\textbf{FID}$\downarrow$}
    & \multirow{2}{*}{\textbf{ID Sim.}$\uparrow$}
    & \textbf{ID Retrieval}$\uparrow$
    & \multirow{2}{*}{\textbf{Pose}$\downarrow$}
    & \multirow{2}{*}{\textbf{Expr.}$\downarrow$} \\
    & & & \textbf{Top-1 / Top-5} & & \\
    \midrule
    SimSwap & 6.16 & 0.67 & 97.89 / 99.00 & 1.94 & 0.78 \\
    CSCS & 13.58 & 0.72 & 94.70 / 97.50 & 2.86 & 1.23 \\
    FaceDancer & 5.48 & 0.57 & 90.77 / 98.40 & 2.16 & 0.72 \\
    InfoSwap & 12.62 & 0.62 & 97.20 / 98.90 & 3.74 & 1.72 \\
    E4S & 19.29 & 0.63 & 94.30 / 97.50 & 3.48 & 1.34 \\
    \midrule
    DiffSwap & 9.53 & 0.36 & 10.10 / 57.10 & 2.10 & 0.86 \\
    DiffFace & 15.97 & 0.66 & 98.08 / 99.04 & 2.88 & 1.05 \\
    FaceAdapter & 8.62 & 0.59 & 89.46 / 91.61 & 4.13 & 1.44 \\
    REFace & 12.03 & 0.74 & 98.60 / 99.10 & 3.35 & 1.29 \\
    \midrule
    \textbf{\ourframework (Student)}, $w$ = 1.0 & 3.25 & 0.64 & 97.69 / 98.90 & 1.97 & 0.76 \\
    \textbf{\ourframework (Student)}, $w$ = 1.5 & 4.37 & 0.68 & 98.29 / 98.90 & 2.23 & 0.81 \\
    \textbf{\ourframework (Student)}, $w$ = 2.0 & 5.39 & 0.69 & 98.49 / 99.30 & 2.43 & 0.86 \\
    \bottomrule
    \end{tabular}
}

\label{tab:faceforensicspp_comparison}
\end{table}

\paragrapht{Extended evaluation on CelebA-HQ and FaceForensics++.}
To address the concern that evaluation on FFHQ alone may not fully reflect generalization, we further evaluate \ourframework on CelebA-HQ and FaceForensics++ (FF++). Compared to FFHQ, these datasets present broader variations in image quality, appearance, and real-world artifacts, providing a more challenging benchmark for jointly assessing identity transfer and attribute preservation.

As shown in \cref{tab:celebahq_comparison,tab:faceforensicspp_comparison} and visualized via Pareto frontiers in \cref{fig:eval_additional_dset}, \ourframework consistently remains on or near the Pareto frontier (upper-right) across all datasets. This performance is particularly notable as it is achieved through training solely on VGGFace2-HQ, whereas several competing methods benefit from dataset-specific training on CelebA-HQ. Furthermore, \ourframework demonstrates superior robustness. While baselines such as REFace and CSCS show severe FID degradation on low-quality datasets like FF++, our method maintains high-quality synthesis and identity consistency.

\subsection{Additional qualitative results on FFHQ}
\begin{figure*}[t] 
    \centering
    \includegraphics[width=\textwidth]{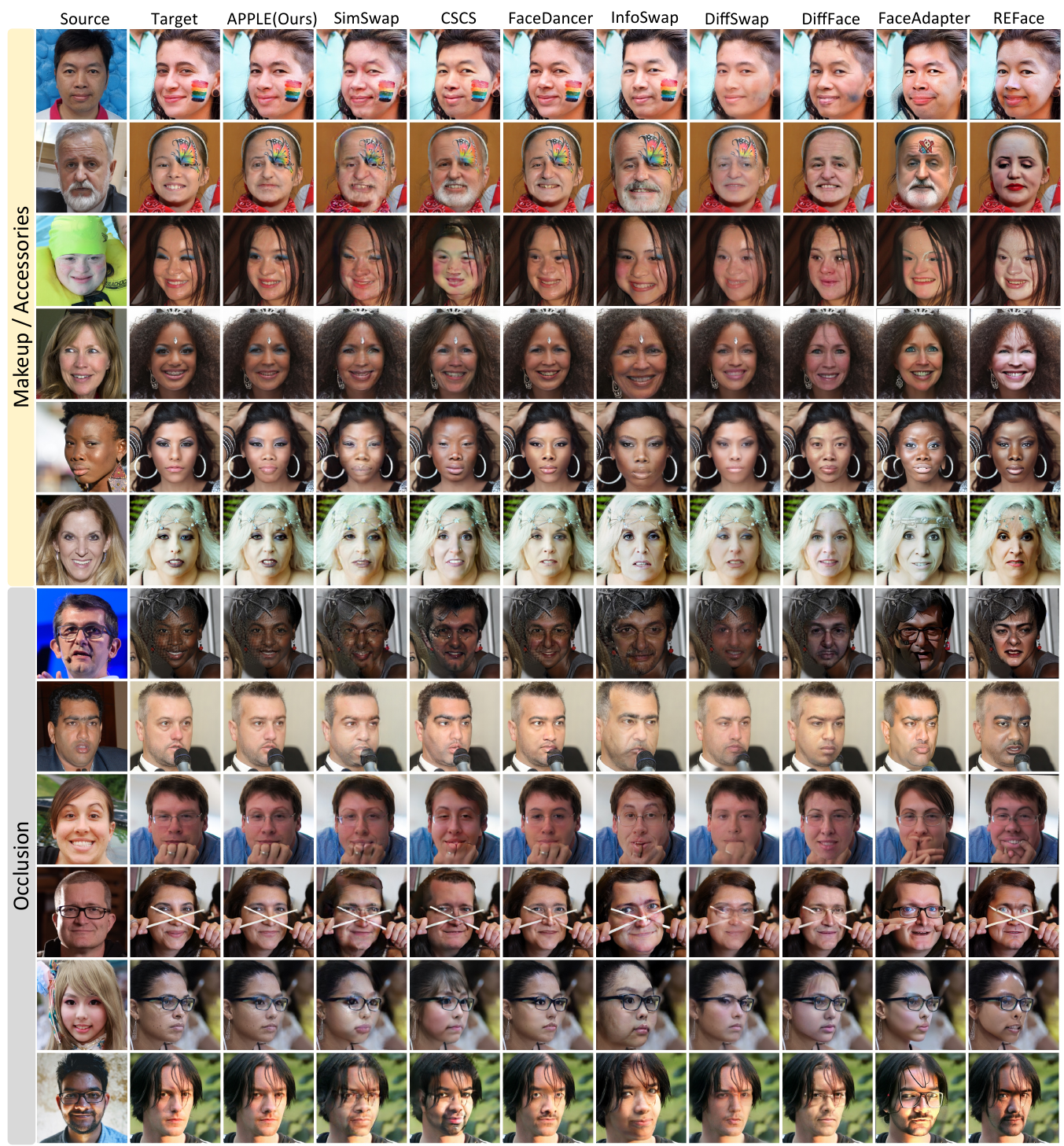}
    \caption{\textbf{Qualitative results.} Face swapping results on the FFHQ dataset~\cite{karras2019style} compared with existing baselines. \ourframework effectively preserves the target image’s attributes while faithfully transferring the source identity.}
    \label{fig:supple_qual_1}
\end{figure*}

\begin{figure*}[t] 
    \centering
    \includegraphics[width=\textwidth]{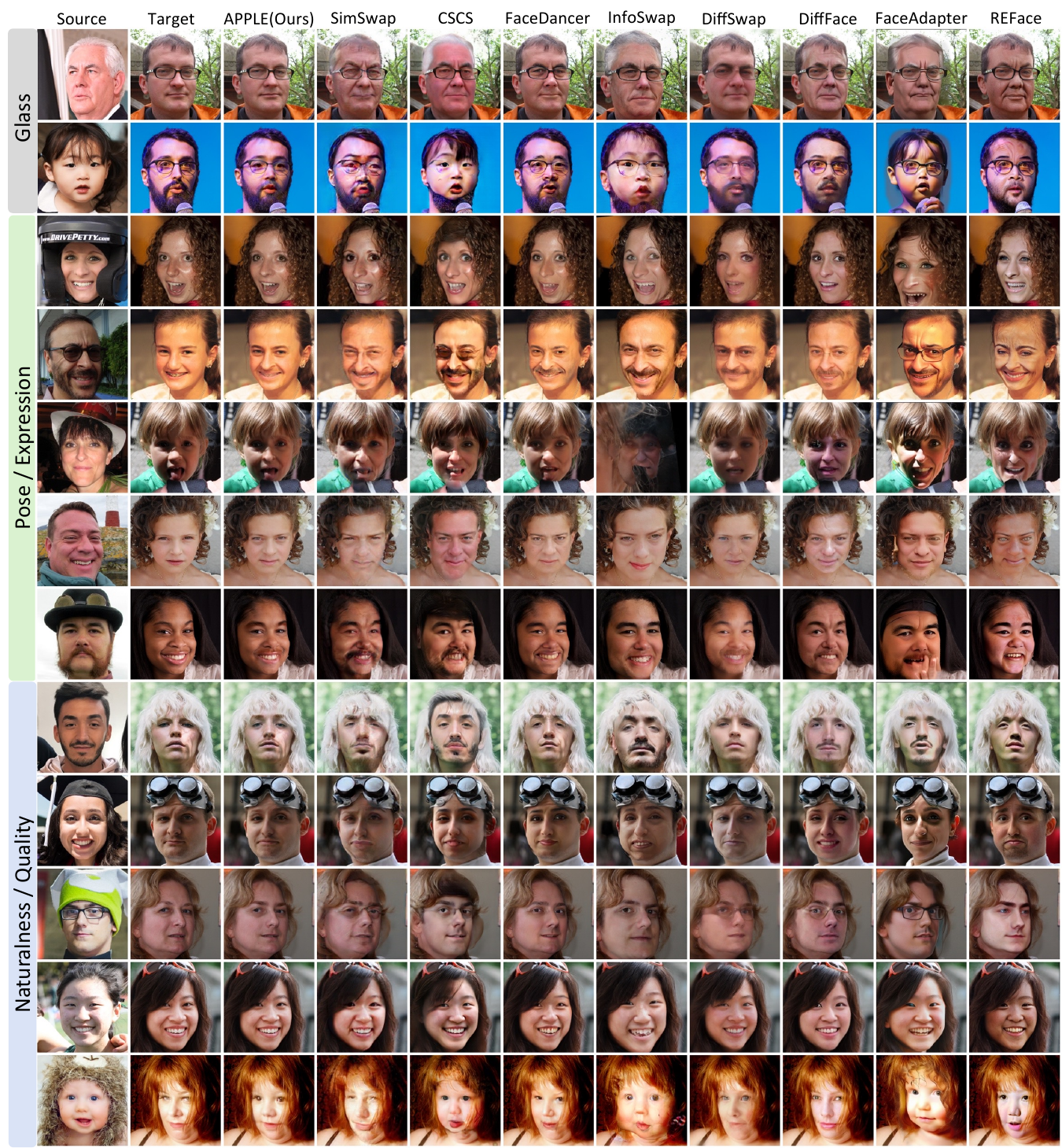}
    \caption{\textbf{Qualitative results.} Face swapping results on the FFHQ dataset~\cite{karras2019style} compared with existing baselines. \ourframework effectively preserves the target image’s attributes while faithfully transferring the source identity.}
    \label{fig:supple_qual_2}
\end{figure*}

\begin{figure*}[t] 
    \centering
    \includegraphics[width=\textwidth]{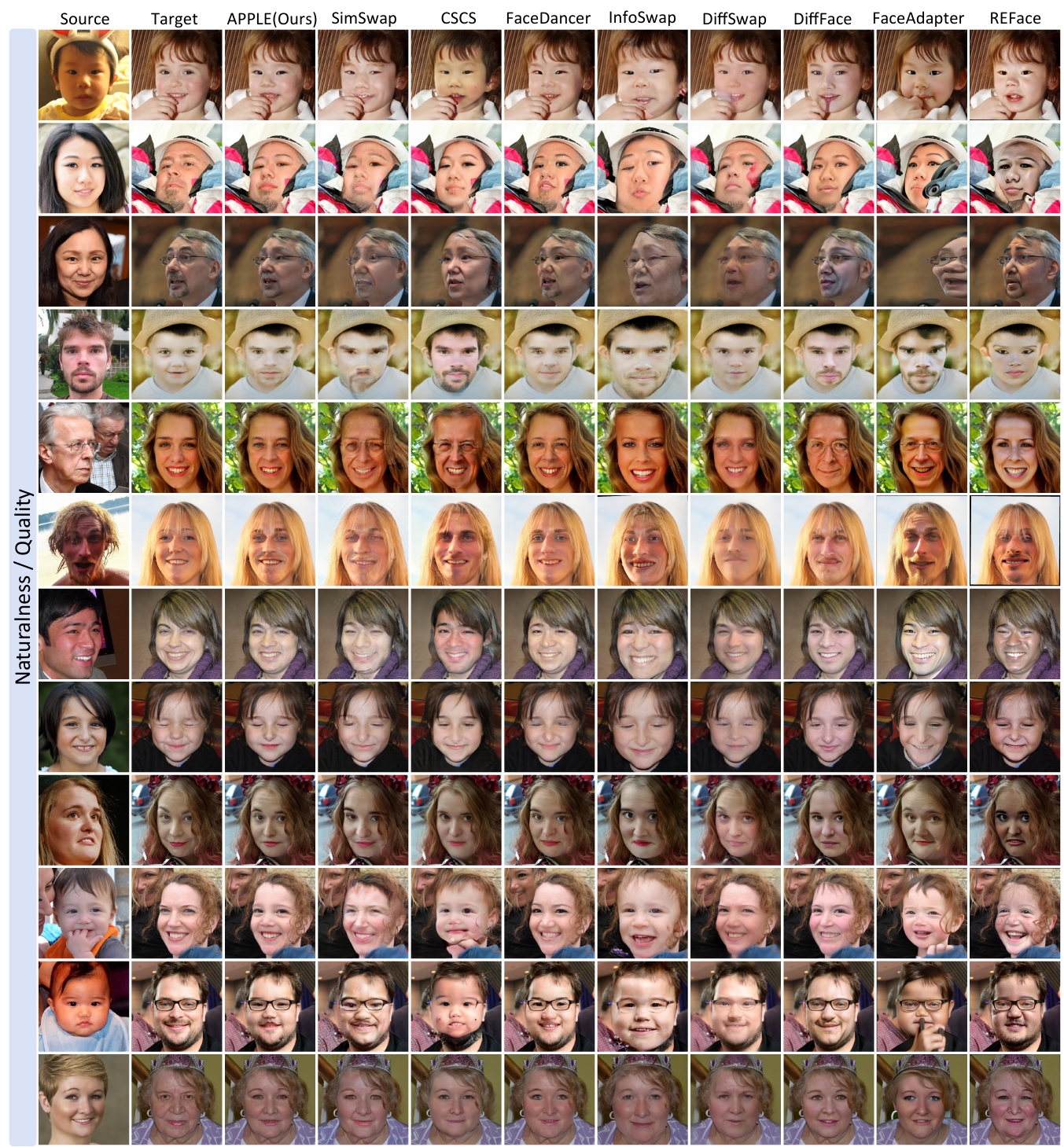}
    \caption{\textbf{Qualitative results.} Face swapping results on the FFHQ dataset~\cite{karras2019style} compared with existing baselines. \ourframework effectively preserves the target image’s attributes while faithfully transferring the source identity.}
    \label{fig:supple_qual_3}
\end{figure*}

\begin{figure*}[t] 
    \centering
    \includegraphics[width=\textwidth]{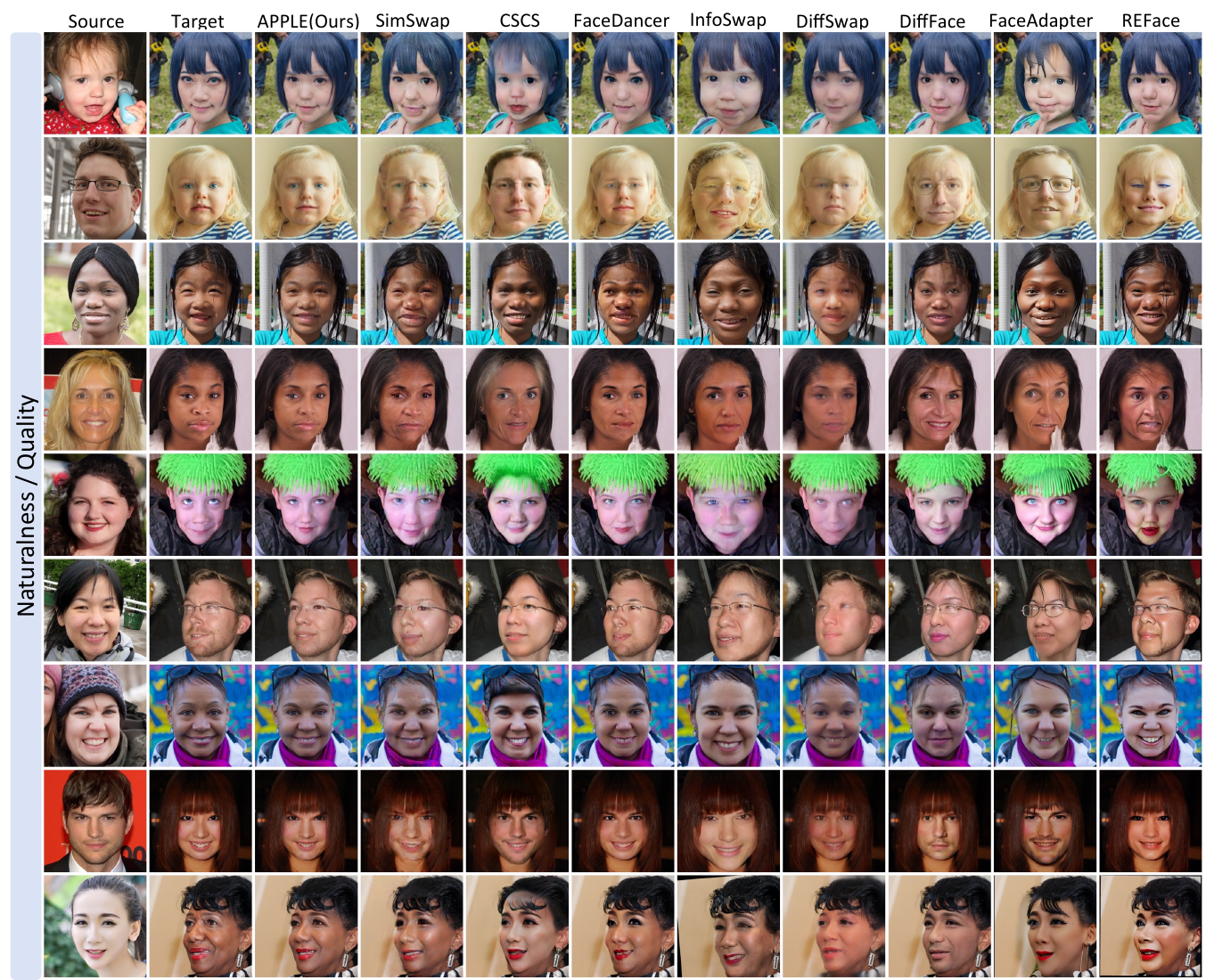}
    \caption{\textbf{Qualitative results.} Face swapping results on the FFHQ dataset~\cite{karras2019style} compared with existing baselines. \ourframework effectively preserves the target image’s attributes while faithfully transferring the source identity.}
    \label{fig:supple_qual_4}
\end{figure*}

We present additional qualitative results of \ourframework (Student) in Fig.~\ref{fig:supple_qual_1}--\ref{fig:supple_qual_4}.

\section{Discussion and Analysis}

\subsection{Detailed motivation and ablation study of conditional deblurring formulation}
\label{supple:ablation_deblur}

In Sec.~\ref{subsec:teacher}, we introduce conditional deblurring as a replacement for the conventional masking strategy used in diffusion-based face swapping. This design choice arises from a fundamental constraint of face swapping task: ground-truth face-swapped pairs do not exist in practice.

\paragrapht{Self-supervised training in GAN-based methods.} GAN-based face swapping models~\citep{li2019faceshifter,chen2020simswap,rosberg2023facedancer} sidestep this inherent limitation by self-supervised training with pairs of images from different identities. Given a source image $I_{\text{src}}^{A}$ of identity $A$ and a target image $I_{\text{tgt}}^{B}$ of identity $B$, the generator is trained with two explicit objectives applied to the generated result $I_{\text{result}}$. An identity loss encourages $I_{\text{result}}$ to share identity with $I_{\text{src}}^{A}$, typically by maximizing the cosine similarity between features extracted from a pretrained identity encoder (e.g., ArcFace~\citep{deng2019arcface}) applied to both images. An attribute-related loss encourages $I_{\text{result}}$ to preserve appearance attributes of $I_{\text{tgt}}^{B}$, commonly implemented by aligning attribute-related features from a custom attribute extractor~\citep{li2019faceshifter}, a GAN discriminator~\citep{chen2020simswap}, or empirically identified attribute-sensitive layers of the identity encoder itself~\citep{rosberg2023facedancer}. Because a GAN produces a clean image in a single forward pass, these feature-based losses can be applied directly to the final output, enabling self-supervised learning without real swapped pairs.

\paragrapht{Why self-supervised paradigm of GAN is inadequate for diffusion models?} This self-supervised paradigm widely used in GAN methods cannot be directly applied to diffusion-based face swapping. Identity and attribute-related features are typically extracted from clean images. During training, the diffusion model predicts $z_0$ from noisy image $z_t$ at various noise levels. At high noise levels (early timesteps), the predicted $z_0$ is far from clean, making feature-based identity and attribute-related losses unreliable or undefined, which leads to training infeasible. While such losses could in principle be applied at low noise levels, this would require excluding high noise levels (early timesteps) from training, severely limiting the model's ability to denoise from initial gaussian noise.


\paragrapht{The proxy formulation in diffusion-based methods.} To address this, existing diffusion-based face swapping approaches adopt inpainting as a proxy training task . Instead of using images from different identities, they use same-identity pairs: a source image $I_{\text{src}}^{A}$ and a target image $I_{\text{tgt}}^{A}$ of the same person $A$. The target image is degraded in a way that suppresses its identity information, while the source image provides identity features from the same person. The diffusion model is then trained to reconstruct the original clean target $I_{\text{tgt}}^{A}$ from its degraded version, conditioned on features from $I_{\text{src}}^{A}$. At inference time, replacing the source with an image from a different identity $B$, while keeping the target from identity $A$, yields a face-swapped result.

In this proxy formulation, the degradation applied to the target image is the mechanism determining how much identity information is removed and how much attribute information remains. Prior works~\citep{zhao2023diffswap,yu2025reface,han2024faceadapter} typically employ full-face masking, replacing the facial region with black image to ensure strong identity suppression. However, such masking discards most attribute cues including lighting, skin tone, makeup, gaze, and expression—which forces the diffusion model to 'imagine' these details during generation. This leads to severe attribute misalignment at inference time, even when identity transfer succeeds.

\paragrapht{Key design questions.} Once we recognize that masking is an unsuitable degradation strategy, adopting an alternative degradation approach requires addressing two key considerations. First, what degradation operator should be applied to the target face in order to reliably suppress identity information. Second, how strong should this degradation be so that identity cues are removed while attribute cues such as lighting, pose, expression, makeup, and accessories remain exploitable by the model.
The degradation design must balance two competing objectives. If the degradation is too weak, identity information from the target is still visible, and the diffusion model can solve the proxy task by reconstructing the target identity, rather than by learning to rely on the source identity. This results in poor identity transfer when the source and target identities differ at inference time. If the degradation is too strong, most appearance cues are destroyed and the model observes little about the target's attributes, which leads it to hallucinate lighting, expression, or makeup in a target-agnostic manner. In that case, even if identity transfer is successful, attribute preservation deteriorates. 

\begin{figure}
    \centering
    \includegraphics[width=\linewidth]{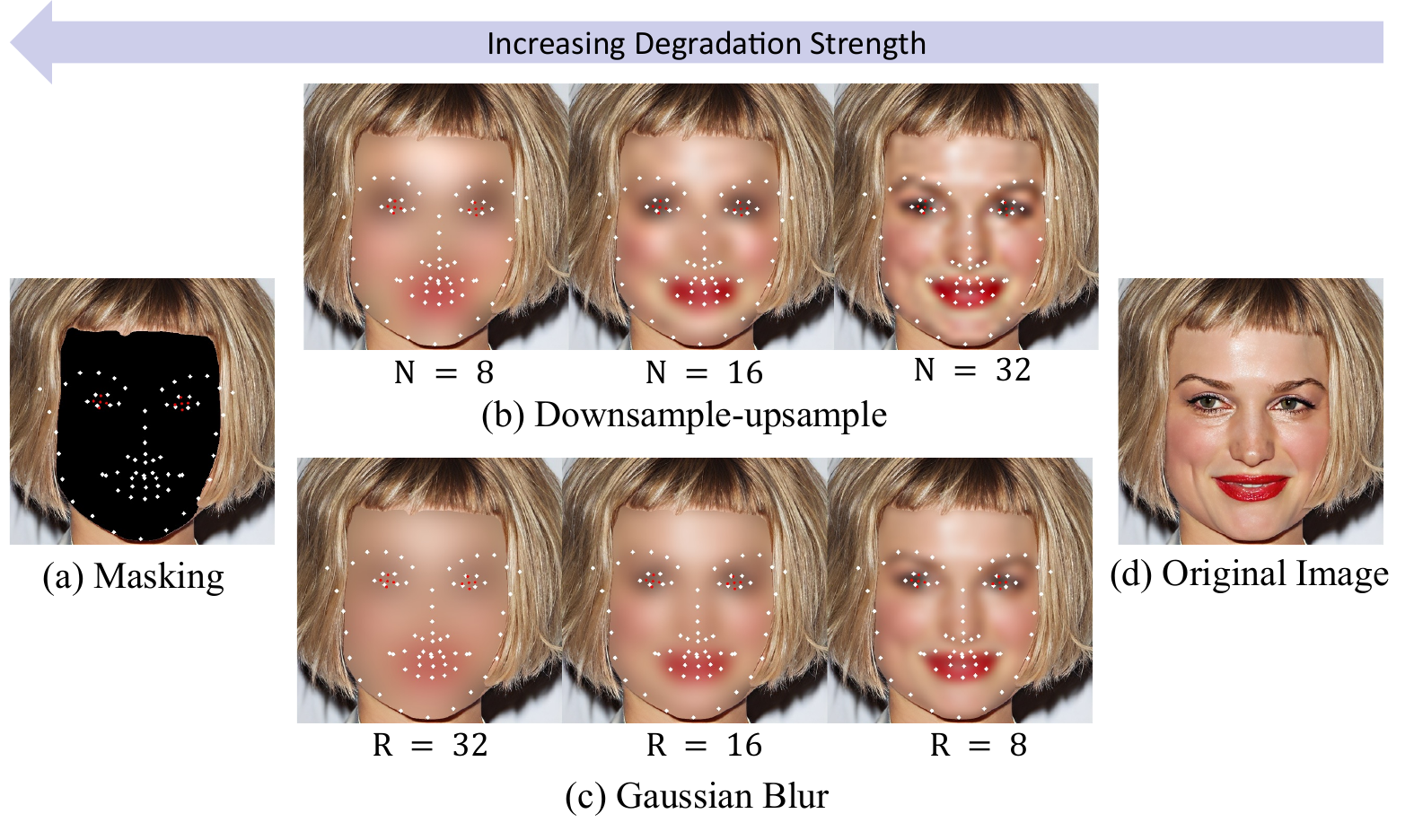}
    \caption{\textbf{Example of considered degradation types and strengths for ablation study of conditional deblurring formulation.}
    }
    \label{fig:example_degradation}
    \vspace{-10pt}
\end{figure}

\begin{figure}
    \centering
    \includegraphics[width=\linewidth]{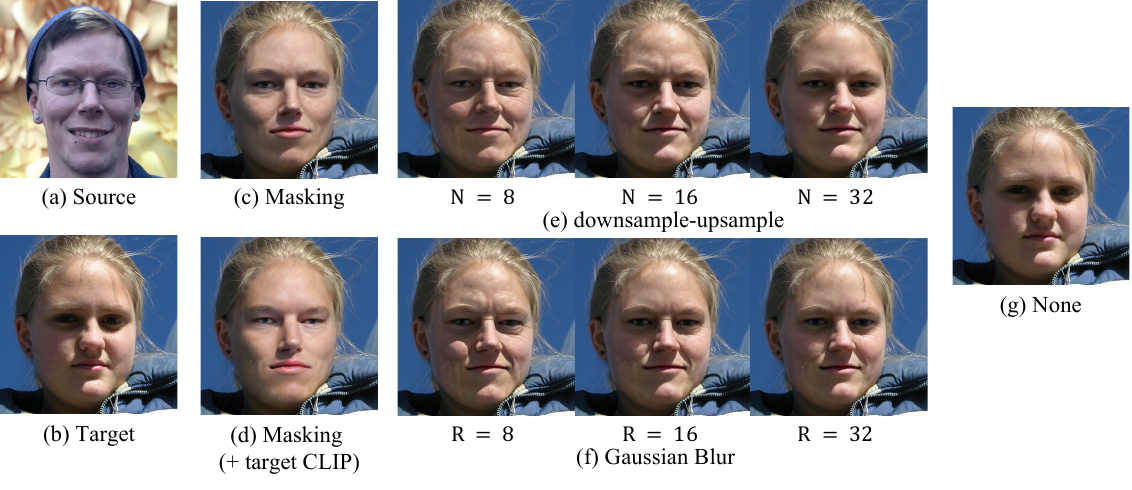}
    \caption{\textbf{Qualitative results of ablation study on conditional deblurring formulation.} 
    `Masking' fails to preserve target attributes (e.g., lighting) due to excessive degradation, even when target CLIP~\citep{radford2021learning_clip} features are provided. Without degradation (`None'), the model simply copies the target image, failing to transfer identity. Moderate degradation (`Downsample-Upsample' or `Gaussian Blur') successfully alters identity while preserving target attributes. Note that stronger degradation improves identity transferability at the cost of precise attribute preservation.
    }
    \label{fig:ablation_deblur_qual}
    \vspace{-10pt}
\end{figure}

\paragrapht{Ablation study design.} To validate our design choices and systematically compare alternatives, we conduct ablation experiments across two dimensions: (1) degradation type and (2) degradation strength. Natural candidates for degradation type include: (1) (fully) masking (2) downsampling--upsampling, (3) Gaussian blurring, and (4) None, as shown in Fig.~\ref{fig:example_degradation}. For downsampling--upsampling and Gaussian blurring, we explore multiple degradation strengths calibrated so that the resulting degraded faces appear qualitatively comparable in terms of visual severity.  All teacher models are trained under a unified protocol for fair comparison: we use a two-phase schedule (training without identity loss followed by training with identity loss), fixing the global step to 5K for each phase, with an effective batch size of 8. In our notation, \texttt{Downsample-\{N\}} refers to downsampling the target face to $N \times N$ resolution followed by upsampling back to the original size. \texttt{GaussianBlur-\{R\}} denotes Gaussian blur applied using \texttt{PIL.ImageFilter.GaussianBlur} with \texttt{radius=R}. We report quantitative results in Tab.~\ref{tab:ablation_deblur_quan} and qualitative results in Fig.~\ref{fig:ablation_deblur_qual}.

The results reveal a clear trend. As the degradation strength increases, identity leakage from the target is reduced, which helps the model rely more consistently on the source identity. This yields higher identity similarity and stronger identity retrieval performance at inference time. In contrast, using a clean, non-degraded target as the conditional input leads to extremely low identity similarity, confirming that explicit identity suppression is necessary.


Increasing degradation strength introduces a moderate drop in pose and expression performance, reflecting the inherent trade-off between identity suppression and attribute preservation. Nonetheless, these metrics remain consistently higher than those obtained with masking-based baselines. At first glance, pose and expression scores may appear only marginally different across degradation types. However, it is important to note that these geometric cues are already provided to the model through landmark conditioning. Qualitative results in Fig.~\ref{fig:ablation_deblur_qual} further demonstrate that both downsampling–upsampling and Gaussian blur preserve a wider range of target attributes (particularly lighting) far more faithfully than masking. This observation is reinforced by FID, which captures overall visual realism: masking-based baselines exhibit substantially worse FID scores because they must hallucinate attributes that are otherwise preserved when using downsampling–upsampling or Gaussian blur degradations.

We also find that, for a matched degradation level, the choice between downsampling--upsampling and Gaussian blur has a smaller impact than the overall strength of degradation itself. Both operators achieve similar identity suppression when calibrated to comparable visual degradation.

Additionally, since some diffusion-based face swapping methods~\citep{yu2025reface,han2024faceadapter} inject target CLIP features as an auxiliary attribute cue, we also evaluate a masking variant conditioned on target CLIP features. However, even with CLIP conditioning, masking fails to adequately preserve target attributes, as shown in Fig.~\ref{fig:ablation_deblur_qual}.

In practice, we adopt the downsampling--upsampling strategy at $8\times8$ resolution as our default. This choice offers a favorable balance between identity suppression and attribute preservation while being simple, stable, and computationally efficient to implement.

\begin{table}[!h]
\centering

\caption{\textbf{Ablation results of conditional deblurring formulation.}
We compare multiple degradation types and strengths used for conditional deblurring. Both downsample–upsample and Gaussian blur exhibit a consistent trend: stronger degradation increases identity suppression, improving identity transfer while weakening attribute preservation. Nevertheless, either degradation method achieves a substantially better identity–attribute balance than masking-based degradation or no degradation at all.
}
\vspace{-5pt}

\resizebox{\linewidth}{!}{
    \begin{tabular}{lccccc}
    \toprule
    \textbf{Degradation type \& strength} & \textbf{FID}$\downarrow$ & \textbf{ID Sim.}$\uparrow$ & \textbf{ID Retrieval}$\uparrow$ & \textbf{Pose}$\downarrow$ & \textbf{Expr.}$\downarrow$ \\
    \midrule
    
    Masking & 7.93 & 0.51 & 87.40 / 95.90 & 2.59 &  0.88 \\
    Masking (+ target CLIP~\citep{radford2021learning_clip}) & 7.87 & 0.42 & 68.10 / 83.60 & 2.47 &  0.83 \\
    
    \midrule
    
    Downsample-8 &  3.86 & 0.51 & 87.40 / 94.20 & 2.42 &  0.78 \\
    Downsample-16 & 2.68 & 0.47 & 80.00 / 90.70 & 2.04 & 0.71 \\
    Downsample-32 & 2.16 & 0.44 & 69.50 / 85.40 & 1.71 & 0.64 \\

    \midrule
    
    GaussianBlur-32 & 3.60 & 0.48 & 83.70 / 91.70 & 2.28 & 0.76 \\
    GaussianBlur-16 & 2.62 & 0.46 & 79.80 / 90.50 & 1.95 & 0.70 \\
    GaussianBlur-8  & 2.08 & 0.45 & 74.50 / 87.90 & 1.72 & 0.65 \\
    
    \midrule
    None & 0.19 & 0.07 & 0.10 / 0.50 & 0.27 & 0.18 \\
    
    \bottomrule
    \end{tabular}
}


\label{tab:ablation_deblur_quan}
\end{table}



\subsection{Design choices for identity loss}
As discussed in Sec.~\ref{supple:ablation_deblur}, applying identity loss directly in diffusion models is nontrivial because they do not produce clean images in a single forward pass. Standard practice in face swapping is to use a pretrained identity embedding encoder such as ArcFace~\citep{deng2019arcface} to enforce similarity between the source identity embedding and the embedding of the generated result. However, these pretrained encoders are typically trained on clean images and are not designed to handle the noisy intermediate predictions produced during diffusion training.

\paragrapht{Existing approaches to identity loss in diffusion models.} Prior diffusion-based face swapping methods~\citep{zhao2023diffswap,yu2025reface} employ various strategies to enable identity supervision. DiffSwap~\citep{zhao2023diffswap} introduces midpoint estimation, in which the model takes two denoising steps from the current noisy latent to obtain a cleaner prediction $z_0$, computes the identity loss on this prediction, and backpropagates the gradient. REFace~\citep{yu2025reface} applies identity loss to the final output $z_0$ obtained after running $N$ DDIM steps from an initial noise. While these approaches enable identity supervision throughout training, they incur substantial memory overhead due to the additional denoising steps required at each training iteration. This makes them difficult to scale to large modern models such as FLUX.1~\citep{flux2024}.

\paragrapht{Our approach.} Following PortraitBooth~\citep{peng2024portraitbooth}, we adopt a simpler strategy: applying identity loss only at low noise levels (late timesteps), where the predicted $z_0$ is sufficiently clean for reliable feature extraction. The key hyperparameter in this approach is the threshold that defines the low noise level. We conduct an ablation study on both teacher and student models to tune this hyperparameter, fixing the training steps to 5K for all configurations. Note that for the teacher model, we initialize from a checkpoint pretrained on FFHQ~\citep{karras2019style} without identity loss, which differs from the teacher setting used in the main paper. The student model configuration remains identical.

For notation, we use \texttt{model(num\%)}, where \texttt{num} denotes the percentage of late timesteps (low noise levels) at which identity loss is applied. For example, Teacher (30\%) indicates that the teacher model was trained with identity loss applied to the latest 30\% of timesteps. As expected, there is a trade-off between identity transferability and attribute preservation. Expanding the range of timesteps at which identity loss is applied improves identity similarity but introduces a moderate drop in attribute-related metrics such as pose and expression accuracy. We select a threshold of 35\% for the teacher model and 50\% for the student model, balancing identity transfer and attribute preservation.

\begin{table}[t!]
\centering

\caption{\textbf{Ablation study of identity loss application across different timestep ranges.} Expanding the range of timesteps at which identity loss is applied improves identity transferability but reduces attribute preservation.
}

\vspace{-5pt}

\resizebox{\linewidth}{!}{
    \begin{tabular}{lccccc}
    \toprule
    \textbf{Method}  & \textbf{FID}$\downarrow$ & \textbf{ID Sim.}$\uparrow$ & \textbf{ID Retrieval}$\uparrow$ & \textbf{Pose}$\downarrow$ & \textbf{Expr.}$\downarrow$ \\
    \midrule
    
    Teacher (30\%) & 4.03& 0.47& 81.70 / 92.50& 2.57&  0.79\\
    Teacher (35\%) & 4.12& 0.48& 83.50 / 94.20& 2.55&  0.80\\

    Teacher (40\%) &  6.26& 0.57& 94.60 / 97.90& 2.88&  0.89\\
    Teacher (50\%) & 7.01& 0.58& 94.60 / 98.00& 3.03& 0.93\\

    \midrule
    
    Student (35\%) & 1.92& 0.49& 82.80 / 92.00& 1.81& 0.61\\
    Student (50\%) & 2.47& 0.53& 87.50 / 95.60& 2.07& 0.65\\
    Student (75\%)  & 3.91& 0.58&  94.00 / 98.10& 2.52& 0.74\\

    \bottomrule
    \end{tabular}
}

\vspace{-10pt}

\label{tab:id_quan}
\end{table}

\subsection{Further boosting GAN via training with our occlusion-augmented pseudo-triplet}


Although diffusion models offer superior generative fidelity, GAN-based face swapping remains widely used in practical applications due to its substantially faster inference speed, as shown in Tab.~\ref{tab:gan_speed}. Improving GAN performance therefore remains an important and complementary research direction. 

Beyond improving diffusion-based models within the teacher–student framework, our attribute-aware pseudo triplets can serve as strong supervision for GAN-based face swapping models also. While some previous works~\cite{huang2024identity_cscs, yuan2023reliableswap} rely on proxy data generated by older face-swapping models~\cite{gao2021information_infoswap} or reenactment systems~\cite{wang2022latent} to compensate for the lack of real supervision, such proxy pairs frequently suffer from noise-like artifacts, inconsistent lighting or gaze, and weak attribute alignment, which ultimately reduces their usefulness as training signals. In contrast, our pseudo triplets preserve target attributes with substantially higher fidelity, providing both qualitatively and quantitatively superior supervision compared to GAN-generated proxies.

\begin{figure}
    \centering
    \includegraphics[width=\linewidth]{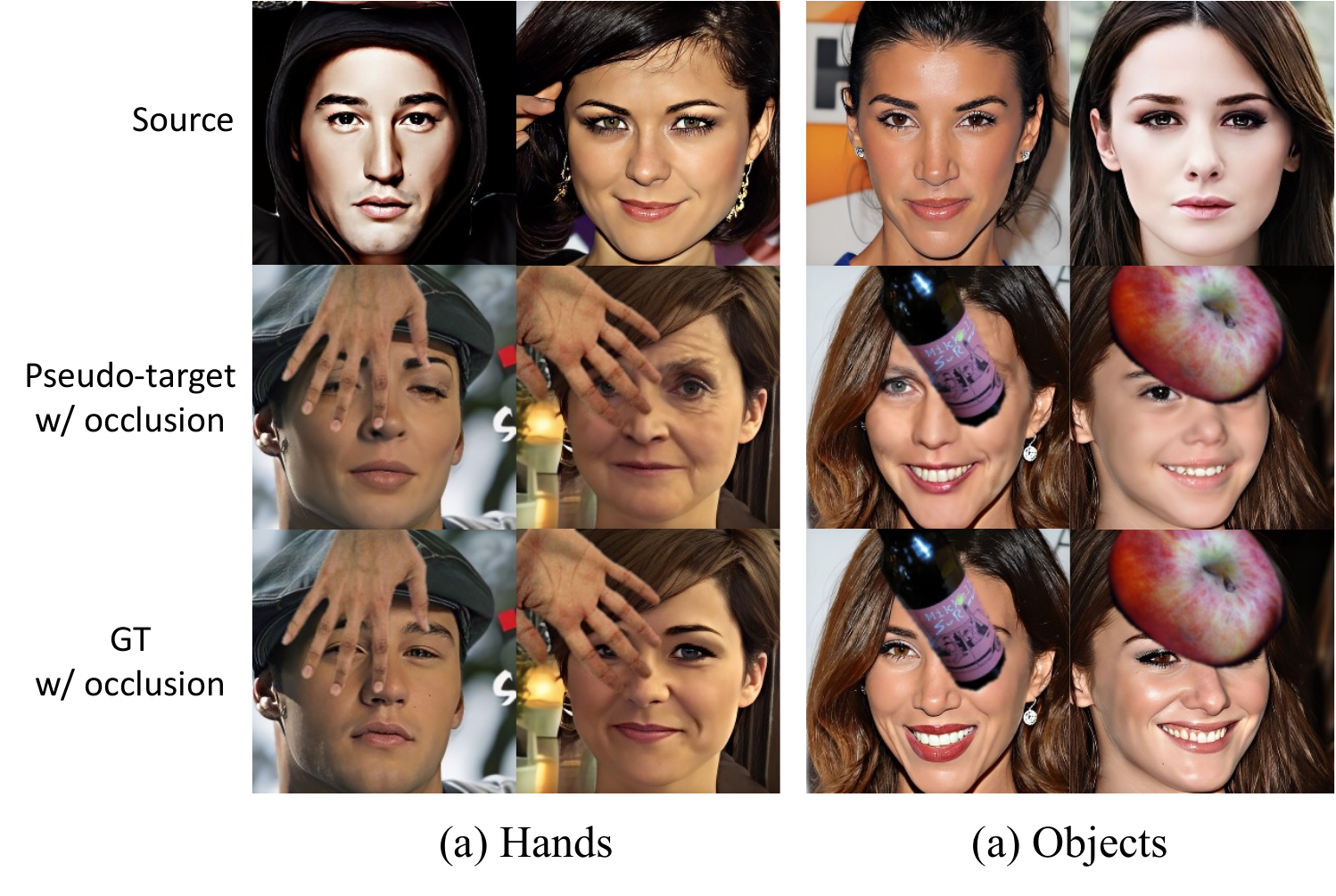}
    \vspace{-20pt}
    \caption{\textbf{Example of applying occlusion augmentation in pseudo-triplets.}}
    \label{fig:occlusion_data}
    \vspace{-10pt}
\end{figure}

\begin{figure}
    \centering
    \includegraphics[width=\linewidth]{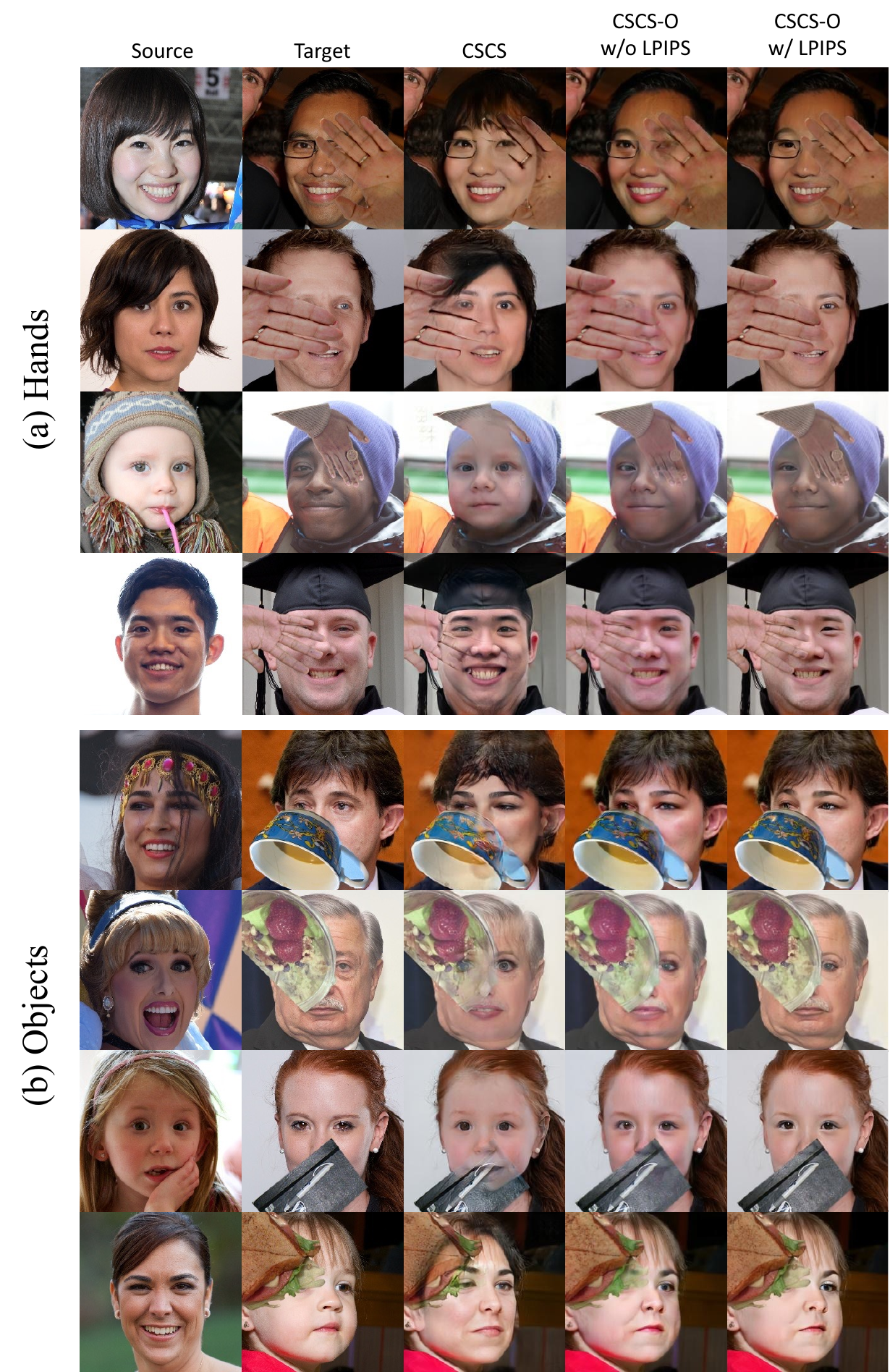}
    \caption{\textbf{Results of CSCS trained with our occlusion-augmented pseudo-triplets.} CSCS-O, which is variant of CSCS~\citep{huang2024identity_cscs} trained with our occlusion-augmented pseudo-triplets, exhibits greatly improved robustness to occlusions compared to baseline (CSCS), especially when trained with LPIPS~\citep{zhang2018perceptual} loss.}
    \label{fig:cscs_qual}
\end{figure}

\begin{table}[!t]
\centering

\caption{\textbf{Comparison of inference speed between GAN and Diffusion model.} Speeds are evaluated on RTX A6000.}
\vspace{-5pt}

    \begin{tabular}{lc}
    \toprule
    \textbf{Model} & \textbf{Second/Sample $\downarrow$} \\
    \midrule
    GAN (CSCS-O) & 0.011    \\
    Diffusion (\ourframework) & 19.858 \\
    \bottomrule
    \end{tabular}


\label{tab:gan_speed}
\end{table}

\begin{table}[h]
\centering

\caption{\textbf{Quantitative results of CSCS-O on FFHQ, which is trained with occlusion-augmented pseudo-triplets generated by \ourframework(Teacher).}
Original CSCS~\citep{huang2024identity_cscs} reports abnormally high identity similarity and retrieval scores due to its copy–paste behavior, which comes at the cost of substantially degraded FID, pose, and expression performance. CSCS-O, which is trained with our occlusion-augmented pseudo-triplets, exhibits markedly stronger robustness to occlusion while maintaining high identity transferability, attribute preservation, and visual fidelity. This demonstrates that our pseudo-triplets provide effective training signals not only for diffusion models but also for GAN-based face swapping.
}
\vspace{-5pt}

\resizebox{\linewidth}{!}{
    \begin{tabular}{lccccc}
    \toprule
    \textbf{Model} & \textbf{FID}$\downarrow$ & \textbf{ID Sim.}$\uparrow$ & \textbf{ID Retrieval}$\uparrow$ & \textbf{Pose}$\downarrow$ & \textbf{Expr.}$\downarrow$ \\
    \midrule
    SimSwap~\cite{chen2020simswap} & 18.54 & 0.55 & 94.10/ 99.00  & 3.11 &  1.73   \\
    MegaFS~\cite{zhu2021one_megafs} & 12.83 & 0.49 & 79.6 / 86.3 & 4.40 &  1.11 \\
    FaceDancer~\cite{rosberg2023facedancer} & 3.80 & 0.51 & 89.70 / 96.50 & 2.23 & 0.74 \\
    HiFiFace~\cite{wang2021hififace} & 11.81 & 0.50 & 85.40 / 93.40 & 3.20 & 1.34 \\
    InfoSwap~\cite{gao2021information_infoswap} & 5.00 & 0.54 & 91.40 / 97.50 & 4.33 & 1.40 \\
    E4S~\cite{liu2023fine_e4s} & 12.13 & 0.49 & 78.30 / 87.80 & 4.39 & 1.29 \\
    \midrule
    DiffSwap~\cite{zhao2023diffswap} & 6.84 & 0.34 & 41.92 / 63.09 & 2.63 & 1.20 \\
    DiffFace~\cite{kim2025diffface} & 8.59 & 0.54 & 90.70 / 95.90 & 3.67 & 1.24 \\
    FaceAdapter~\cite{han2024faceadapter} & 13.03 & 0.52 & 87.00 / 93.20 & 5.12 & 1.38 \\
    REFace~\cite{yu2025reface} & 7.22 & 0.60  &  97.60 / 99.40 & 3.67 & 1.08 \\
    \midrule
    CSCS~\cite{huang2024identity_cscs} & 11.00 & 0.65 & 99.00 / 99.50 & 3.64 &  1.44 \\
    \textbf{CSCS-O (w/o LPIPS) } & 5.98& 0.52& 90.20 / 97.10& 2.04& 1.23\\
    \textbf{CSCS-O (w/ LPIPS) } & 4.55 & 0.54 & 90.40 / 96.40 & 2.12 & 1.30 \\
    \bottomrule
    \end{tabular}
}


\label{tab:gan_quan}
\end{table}

In addition, our pseudo triplets can be further extended with occlusion augmentation, enabling direct supervision for training models to be robust under real-world occlusions such as hands, accessories, or objects covering the face.
As show in Fig.~\ref{fig:occlusion_data}, we overlay hand and random object occlusions on the target images in our pseudo-triplets and use these occluded pairs as supervision, following~\cite{voo2022delving}. A key challenge here is that the identity loss encourages the generator to reconstruct the source identity across the entire facial region, which causes the model to overwrite occluding objects with hallucinated face components. To prevent this undesired behavior, we additionally employ an LPIPS~\citep{zhang2018perceptual} loss between the generated image and the occluded pseudo target. This perceptual loss encourages the generator to respect the structure and texture of the occlusion, preventing it from being incorrectly “corrected” to a facial region. As a result, LPIPS stabilizes fine-tuning under occlusion, enabling the model to maintain identity similarity where appropriate while faithfully preserving the occluding objects.

We evaluate the effectiveness of our occlusion-augmented pseudo-triplet by fine-tuning a representative GAN-based model, CSCS~\citep{huang2024identity_cscs}, and report quantitative results in Tab.~\ref{tab:gan_quan} and qualitative results in Fig.~\ref{fig:cscs_qual}. We denote the model trained with our occlusion-augmented pseudo-triplets as CSCS-O. Qualitatively, CSCS-O demonstrates markedly improved robustness to occlusions compared to models trained with conventional proxy data, particularly in settings where LPIPS loss is used. Quantitatively, it yields substantial gains in FID, pose, and expression consistency while maintaining strong identity transferability.


\subsection{Training cost breakdown}
\begin{table}[t]
\centering

\caption{\textbf{Training cost breakdown.} Cost is calculated assuming $4\times$RTX A6000 GPUs. The full training pipeline takes approximately 13 days, which is reasonable even compared to single-stage methods. For example, REFace reports 18 days of training cost on $2\times$A100 GPUs.}
\vspace{-5pt}

\begin{tabular}{lc}
    \toprule
    Stage & Time (days) \\
    \midrule
    Teacher training (65K steps) & 4.3 \\
    Pseudo-triplet generation (50K images) & 5.78 \\
    Student training (15K steps) & 2.4 \\
    \midrule
    \textbf{Total} & \textbf{12.78} \\
    \bottomrule
\end{tabular}

\label{tab:training_cost_breakdown}
\end{table}

We report the training cost breakdown of our teacher-student framework in Tab.~\ref{tab:training_cost_breakdown}. The teacher model is trained for 15K iterations without identity loss and 50K iterations with identity loss, while the student model is trained for 15K iterations with pseudo-triplet generated by the teacher model. The overall training time is approximately 13 days on 4$\times$NVIDIA A6000 GPUs.

\section{Limitations and future directions}
While our teacher–student framework generally performs robustly, it still relies on several external modules such as 3DMM landmarks, gaze landmarks, and segmentation maps, when constructing pseudo-triplets for student training. Errors introduced by these modules may propagate into the pseudo-triplet, which may influence the downstream student training. Reducing this dependency, or designing architectures that are more resilient to errors of external module, is an important direction for future work. Additionally, improving the inference efficiency of diffusion models and extending the framework to video face swapping present promising avenues for further research.

\FloatBarrier
\clearpage



\end{document}